\documentclass[letterpaper]{article} % DO NOT CHANGE THIS
\usepackage{aaai23}  % DO NOT CHANGE THIS
\usepackage{times}  % DO NOT CHANGE THIS
\usepackage{helvet}  % DO NOT CHANGE THIS
\usepackage{courier}  % DO NOT CHANGE THIS
\usepackage[hyphens]{url}  % DO NOT CHANGE THIS
\usepackage{graphicx} % DO NOT CHANGE THIS
\usepackage{natbib}  % DO NOT CHANGE THIS AND DO NOT ADD ANY OPTIONS TO IT
\usepackage{caption} % DO NOT CHANGE THIS AND DO NOT ADD ANY OPTIONS TO IT
\newcommand{\algo}{\texttt{KSRL}} 

\usepackage{algorithm}
\usepackage{algorithmic}

\usepackage{newfloat}
\usepackage{listings}
\DeclareCaptionStyle{ruled}{labelfont=normalfont,labelsep=colon,strut=off} % DO NOT CHANGE THIS
\floatstyle{ruled}
\newfloat{listing}{tb}{lst}{}
\floatname{listing}{Listing}
\usepackage{microtype}
\usepackage{amssymb,amsmath, amsthm,amsthm}
\usepackage[capitalize]{cleveref} 
\usepackage{graphicx}
 \usepackage{color}
\usepackage{caption}
\usepackage{subcaption}
\usepackage{wrapfig, blindtext}
\usepackage{booktabs} % for professional tables
\usepackage{amssymb}
\usepackage{framed}
\usepackage{mathrsfs}
\usepackage{multirow}
\usepackage{enumerate,pifont,bigdelim,bbding}
\usepackage{amsthm}

\newtheorem{theorem}{Theorem}[section]

\newtheorem{lemma}[theorem]{Lemma}

\newcommand{\ip}[1] {\langle #1 \rangle }

\newcommand{\inclu}[0] {\ar@{^{(}->}}

\title{Posterior Coreset Construction with Kernelized Stein Discrepancy\\ for Model-Based Reinforcement Learning}
\author{
Souradip Chakraborty\textsuperscript{\rm 1}, 
Amrit Singh Bedi\textsuperscript{\rm 1},
Alec Koppel\textsuperscript{\rm 2},\\
Pratap Tokekar\textsuperscript{\rm 1},
Brian Sadler\textsuperscript{\rm 3},
Furong Huang\textsuperscript{\rm 1},
Dinesh Manocha\textsuperscript{\rm 1}
}
\affiliations{
    \textsuperscript{\rm 1}University of Maryland, College Park, USA, \\
    \textsuperscript{\rm 2}JP Morgan AI Research, NY, USA,\\ \textsuperscript{\rm 3}DEVCOM Army Research Laboratory, Adeplhi, USA\\
	\{schakra3,amritbd,tokekar, furongh, dmanocha\}@umd.edu, alec.koppel@jpmchase.com, brian.m.sadler6.civ@army.mil
}

\usepackage{bibentry}
\begin{document}

\maketitle

\begin{abstract}
Model-based approaches to reinforcement learning (MBRL) exhibit favorable performance in practice, but their theoretical guarantees in large spaces are mostly restricted to the setting when transition model is Gaussian or Lipschitz, and demands a posterior estimate whose representational complexity grows unbounded with time. In this work, we develop a novel MBRL method (i) which relaxes the assumptions on the target transition model to belong to a generic family of mixture models; (ii) is applicable to large-scale training by incorporating a compression step such that the posterior estimate consists of a Bayesian coreset of only statistically significant past state-action pairs; and (iii) exhibits a sublinear Bayesian regret.
To achieve these results, we adopt an approach based upon Stein's method, which, under a smoothness condition on the constructed posterior and target, allows distributional distance to be evaluated in closed form as the kernelized Stein discrepancy (KSD). The aforementioned compression step is then computed in terms of greedily retaining only those samples which are more than a certain KSD away from the previous model estimate.
Experimentally, we observe that this approach is competitive with several state-of-the-art RL methodologies, and can achieve up-to 50 percent reduction in wall clock time in some continuous control environments. \end{abstract}

\section{Introduction}
Reinforcement learning, mathematically characterized by a Markov Decision Process (MDP) \cite{puterman2014markov}, has gained traction for addressing sequential decision-making problems with long-term incentives and uncertainty in state transitions \cite{sutton1998reinforcement}. A persistent debate exists as to whether model-free (approximate dynamic programming \cite{sutton1988learning} or policy search \cite{williams1992simple}), or model-based (model-predictive control, MPC  \cite{garcia1989model,kamthe2018data}) methods, are superior in principle and practice \cite{wang2019benchmarking}. A major impediment to settling this debate is that performance certificates are presented in disparate ways, such as probably approximate correct (PAC) bounds \cite{strehl2009reinforcement,dann2017unifying}, frequentist regret \cite{jin2018q,jin2020provably}, Bayesian regret \cite{agrawal2017optimistic,xu2020reinforcement,o2021variational}, and convergence in various distributional metrics \cite{borkar2002risk,amortila2020distributional,kose2021risk}. In this work, we restrict focus to regret, as it imposes the fewest requirements on access to a generative model underlying state transitions.

%paragraph about frequentist regret bounds, upper confidence bound 
In evaluating the landscape of frequentist regret bounds for RL methods, both model-based and model-free approaches have been extensively studied \cite{jin2018q}. Value-based methods in episodic settings have been shown to achieve regret bounds $\tilde{\mathcal{O}}(d^p H^q \sqrt{T})$ \cite{yang2020reinforcement} (with $p=1$, $q=2$), where $H$ is the episode length, and $d$ is the aggregate dimension of the state and action space. This result has been improved to $p=q=3/2$ in \cite{jin2020provably}, and further to $p=3/2$ and $q=1$ in \cite{zanette2020learning}. Recently, model-based methods have gained traction for improving upon the best known regret model-free methods with $p=1$ and $q=1/3$ \cite{ayoub2020model}. A separate line of works seek to improve the dependence on $T$ to be logarithmic through instance dependence \cite{JMLR:v11:jaksch10a,zhou2021nearly}. These results typically impose that the underlying MDP has a transition model that is linearly factorizable, and exhibit regret depends on the input dimension $d$. This condition can be relaxed through introduction of a nonlinear feature map, whose appropriate selection is highly nontrivial and lead to large gaps between regret and practice \cite{nguyen2013online}, or meta-procedures \cite{lee2021online}. Aside from the feature selection issue, these approaches require evaluation of confidence sets which is computationally costly and lead to statistical inefficiencies when  approximated \cite{osband2017posterior}.

Thus, we prioritize \emph{Bayesian} approaches to RL \cite{ghavamzadeh2015bayesian,kamthe2018data} popular in robotics  \cite{deisenroth2013gaussian}. While many heuristics exist, performance guarantees take the form of Bayesian regret \cite{https://doi.org/10.48550/arxiv.1306.0940}, and predominantly build upon posterior (Thompson) sampling \cite{thompson1933likelihood}. In particular, beyond the tabular setting, \cite{osband2014model} establishes a $\tilde{O}(\sigma_R\sqrt{d_K(R)d_E(R)T}+\mathbb{E}[L^*]\sigma_p\sqrt{d_K(P)d_E(P)})$  Bayesian regret for posterior sampling RL (PSRL) combined with greedy action selections with respect to the estimated value. Here $L^*$ is a global Lipschitz constant for the future value function, $d_K$ and $d_E$ are Kolmogorov and eluder dimensions, and $R$ and $P$ refers to function classes of rewards and transitions. The connection between $H$ and $L$ is left implicit; however, \cite{pmlr-v139-fan21b}[Sec. 3.2] shows that $L$ can depend exponentially on $H$. Similar drawbacks manifest in the Lipschitz parameter of the Bayesian regret bound of \cite{chowdhury2019online}, which extends the former result to continuous spaces through kernelized feature maps. However, recently an augmentation of PSRL is proposed which employs feature embedding with Gaussian (symmetric distribution) dynamics to alleviate this issue \cite{pmlr-v139-fan21b}, yielding the Bayesian regret of $\tilde{\mathcal{O}}(H^{\frac{3}{2}}d\sqrt{T})$ that is polynomial in $H$ and has no dependence on Lipschitz constant $L$. These results are still restricted in the sense that it requires { (i) the transition model target to be Gaussian, (ii) its representational complexity to grow  unsustainably large with time.} 
Therefore, in this work we as the following question: 

{ \emph{Can we achieve a trade-off between the Bayesian regret and the posterior representational complexity (aka coreset size) without oracle access to a feature map at the outset of training, in possibly continuous state-action spaces?}} 

We provide an affirmative answer by honing in on the total variation norm used to quantify the posterior estimation error that appears in the regret analysis of \cite{pmlr-v139-fan21b}, and identify that it can be sharpened by instead employing an integral probability metric (IPM). Specifically, by shifting to an IPM, and then imposing structural assumptions on the target, that is, restricting it {to a class of smooth densities}, we can employ \emph{Stein's identity} \cite{stein1956inadmissibility,kattumannil2009stein} to evaluate the distributional distance in closed form using the kernelized Stein discrepancy (KSD) \cite{gorham2015measuring,liu2016kernelized}. This restriction is common in Markov Chain Monte Carlo (MCMC) \cite{andrieu2003introduction}, and imposes that the target, for instance, belongs to a family of mixture models. 

This modification in the metric of convergence alone leads to improved regret because we no longer require the assumption that the posterior is Gaussian \cite{pmlr-v139-fan21b}.  However, our goal is to translate the scalability of PSRL from tabular settings to continuous spaces {which requires addressing the parameterization complexity of the posterior estimate, which grows linearly unbounded \cite{pmlr-v139-fan21b} }. With the power to evaluate KSD in closed form, then, we sequentially remove those state-action pairs that contribute least {(decided by a compression budget $\epsilon$)} in KSD after each episode ({which is completely novel in the RL setting}) from the posterior representation according to \cite{hawkins2022online}. Therefore, the posterior estimate only retains statistically significant past samples from the trajectories, i.e., it is defined by a Bayesian coreset of the trajectory data \cite{campbell2018bayesian,campbell2019automated}.  The budget parameter $\epsilon$ then is calibrated in terms of a rate determining factor $\alpha$ to yield both sublinear Bayesian regret and sublinear representational complexity of the learned posterior -- see  Table \ref{table_intro}. 
The resultant procedure we call Kernelized Stein Discrepancy-based Posterior Sampling for RL ({\algo}). Our main contributions are, then, to:
\begin{list}{$\rhd$}{\topsep=0.ex \leftmargin=0.2in \rightmargin=0.in \itemsep =0.0in}
    \item introduce Stein's method in MBRL for the first time, and use it to develop a novel transition model estimate based upon it, which operates in tandem with a KSD compression step to remove statistically insignificant past state-action pairs, which we abbreviate as {\algo};
    \item establish Bayesian regret bounds of the resultant procedure that is sublinear in the number of episodes experienced, without any prior access to a feature map, alleviating difficult feature selection drawbacks of prior art. Notably, these results relax Gaussian and Lipschitz assumptions of prior related results;
    \item {mathematically establish a tunable trade-off between Bayesian regret and posterior’s parameterization complexity (or dictionary size) via introducing parameter $\alpha\in(0,1]$ for the first time in this work;}
    \item experimentally demonstrate that {\algo} achieves favorable tradeoffs between sample and representational complexity relative to several strong benchmarks.
\end{list}

\begin{table*}[t]
\centering
%\resizebox{0.75\textwidth}{!}{
\begin{tabular}{|c|c|c|c|}
\hline
\hline
Setting & \textbf{Refs} & \textbf{Bayes Regret} & \textbf{{Coreset Size}}  \\ \hline\hline
Tabular & PSRL \cite{osband2013}        &  $\tilde{ \mathcal{O}}(HS \sqrt{AT})$        &    $\Omega(T)$               \\ 
\hline
%Lazy PSRL \cite{abbasi2015bayesian}         &        &                 &  Gap in Theorem           \\ \hline
Tabular &PSRL2 \cite{osband17a}          & $\tilde{\mathcal{O}}(H \sqrt{SAT})$       &  $\Omega(T)$                  \\ \hline
Tabular &TSDE \cite{ouyang2017control}      & $\tilde{\mathcal{O}}(HS \sqrt{AT})$       &        $\Omega(T)$               \\ \hline
Tabular &General PSRL \cite{agrawal2017optimistic}        &  $\tilde{\mathcal{O}}(DS \sqrt{AT})$      &          $\Omega(T)$               \\ \hline
Tabular &DS-PSRL \cite{theocharous2018scalar}      & $\tilde{\mathcal{O}}(CH \sqrt{C'T})$       &          $\Omega(T)$                 \\ \hline
Tabular & PSRL3 \cite{osband2014model}          & $\tilde{ \mathcal{O}}(\sqrt{d_k d_E T})$       &            $\Omega(T)$     \\ \hline
Continuous/Gaussian & MPC-PSRL \cite{pmlr-v139-fan21b}          & $\tilde{\mathcal{O}}(H^{\frac{3}{2}}d\sqrt{T})$       &             $\Omega(T)$           \\ \hline
Continuous/ Smooth & \texttt{KSRL} (\textbf{This work})          &   $\tilde{\mathcal{O}}\left(dH^{1+({\alpha}/{2})}T^{1-({\alpha}/{2})}\right)$     &      $\Omega(\sqrt{T^{1+\alpha}})  $             \\ \hline
\end{tabular}
%}
%\vspace{5mm}
\caption{A comparison of Bayes regret (cf. \eqref{bayes_regret}) and {Bayesian Coreset} (the number of stored {data points} in dictionary $\mathcal{D}_k$ to represent posterior at $k$). We introduce KSD-based compression to model-based RL (KSRL), with tuning parameter $\alpha$ to obtain sublinear Bayesian regret \emph{and} {coreset size} for any $\alpha\in(0,1]$. For $\alpha=1$, { we recover the state of the art results of MPC-PSRL ($\tilde{\mathcal{O}}(dH^{3/2}\sqrt{T})$). But our results hold for general transitions, which are smooth and not restricted to Gaussian assumption. }}\label{table_intro}% 
\end{table*}

\section{Problem Formulation} 

We consider the problem of modelling an episodic finite-horizon Markov Decision Process (MDP) where the true unknown MDP is defined as $M^* :=\{\mathcal{S},\mathcal{A},R^*,P^*,H, R_{\text{max}},\rho\}$, where $\mathcal{S}\subset \mathbb{R}^{d_s}$ and $\mathcal{A}\subset \mathbb{R}^{d_a}$ denote continuous state and action spaces, respectively. Here, $P^*$ represents the true underlying  generating process for the state action transitions and $R^*$  is the true rewards distribution. 
After every episode of length $H$, the state will reset according to the initial state distribution  $\rho$. At time step $i \in [1,H]$ within an episode, the agent observe $s_i \in \mathcal{S}$, selects $a_i \in \mathcal{A}$ according to a policy $\mu$, receives a reward  $r_i \sim R^*(s_{i},a_i)$ and transitions to a new state $s_{i+1} \sim P^*(\cdot|s_{i},a_i)$. We consider $M^*$ itself as a random process, as is the often the case in Bayesian Reinforcement Learning,  which helps us to distinguish between the true and fitted transition/reward model.

Next, we define policy $\mu$ as a mapping from state $s \in \mathcal{S}$ to action $a \in \mathcal{A}$ over an episode of length $H$. For a given MDP $M$, the value for time step $i$  is the reward accumulation during the episode:
\begin{align}
    V^{M}_{\mu, i}(s)=\mathbb{E}[\Sigma_{j=i}^{H}[\bar{r}^M(s_{j},{\mu(s_j,j)})|s_i = s],
\end{align}
where actions are under policy $\mu(s_j,j)$ ($j$ denotes the timestep within the episode)  and $\bar{r}^M(s,a) = \mathbb{E}_{r \sim R^M(s,a)} [r]$. Without loss of generality, we assume the expected reward an agent receives at a single step is bounded $|\bar{r}^M(s,a)| \leq R_{\text{max}}$, $\forall s\in \mathcal{S},a\in \mathcal{A}$. This further implies that $|V(s)|\leq HR_{\text{max}}$, $\forall s$. For a given MDP $M$, the optimal policy $\mu^M$ is defined as 
\begin{align}\label{optimal_policy}
    \mu^M = \arg\max_{\mu} V^{M}_{\mu, i}(s),
\end{align}
for all $s$ and $i=1,\dots,H$. Next, we also define future value function $U^{M}_i(P)$ to be the expected value of the value function over all initializations and trajectories
\begin{align}\label{value_function}
    U^{M}_{i}(P)=\mathbb{E}_{s'\sim P(s'|s,{a}), a= \mu^M(s,i)}[ V_{\mu^M, i}^{M} (s')|s_i = s],
\end{align}
where $P$ is the transition distribution under MDP $M$. According to these definitions, one would like to find the optimal policy \eqref{optimal_policy} for the true model $M=M^*$.

Next, we review the PSRL algorithm, which is an adaption of Thompson sampling to RL \cite{osband2014model} (see Algorithm \ref{alg:PSRL} in Appendix \ref{sec:Prelim}).
 In PSRL, we start with a prior distribution over MDP given by $\phi$. Then at each episode, we take sample $M^k$ from the posterior given by $\phi(\cdot|\mathcal{D}_k)$ , where $\mathcal{D}_{k} :=\{s_{1,1},a_{1,1},r_{1,1},\cdots,s_{k-1,H},a_{k-1,H},r_{k-1,H} \}$ is a data set containing past trajectory data, i.e., state-action-reward triples, which we call a \emph{dictionary}. That is, where $s_{k,i},
a_{k, i}$ and $r_{k, i}$ indicate the state, action, and reward at time step $zi$ in episode $k$. Then, we evaluate the optimal policy $\mu^{k}:=\mu^{M^k}$ via \eqref{optimal_policy}. Thereafter, information from the latest episode is appended to the dictionary as  $\mathcal{D}_{k+1}= \mathcal{D}_{k}\cup \{s_{k,1},a_{k,1},r_{k,1},\cdots,s_{k,H},a_{k,H},r_{k,H} \}$. 

\textbf{Bayes Regret and Limitations of PSRL:} To formalize the notion of performance in the model-based RL setting, we define Bayes regret for episode $k$ as \cite{osband2014model,pmlr-v139-fan21b}
\begin{equation}\label{regret_0}
\Delta_k = \int \rho(s_1) (V_{\mu^*,1}^{M^*}(s_1)-V_{\mu^k, 1}^{M^*}(s_1)) ds_1, 
\end{equation}
where $\rho(s_1)$ is the initial state distribution, and $\mu^k$ is the optimal policy for $M^k$ sampled from posterior at episode $k$. The total regret for all the episodes is thus given by 
\begin{align}\label{bayes_regret}
     {Regret}_T:=& \Sigma_{k=1}^{\lceil\frac{T}{H}\rceil} \Delta_k, \ \ \ \text{and} \ \ \ \\ Bayes Regret_T :=& \mathbb{E}[Regret_T~|~ M^*\sim \phi].
\end{align}
The Bayes regret of PSRL (cf. Algorithm \ref{alg:PSRL}) is established to be $\tilde{\mathcal{O}}(\sqrt{d_K(R)d_E(R)T}+\mathbb{E}[L^*]\sqrt{d_K(P)d_E(P)})$ where $d_K$ and $d_E$ are Kolmogorov and Eluder dimensions, $R$ and $P$ refer to function classes of rewards and transitions, and $L^*$ is a global Lipschitz constant for the future value function. Although it is mentioned that system noise smooths the future value functions in \cite{osband2014model}, an explicit connection between $H$ and $L$ is absent, which leads to an exponential dependence on horizon length $H$ in the regret \cite[Corollary 2]{osband2014model} for LQR. This dependence has been improved to a polynomial rate in $H$: $\tilde{\mathcal{O}}(H^{\frac{3}{2}}d\sqrt{T})$ in  \cite{pmlr-v139-fan21b}, which is additionally linear in $d$ and sublinear in $T$. 

{A crucial assumption in deriving the best known regret bound for PSRL with continuous state action space is of target distribution belonging to Gaussian/symmetric class, which is often violated. For instance, if we consider a variant of inverted pendulum with an articulated arm, the transition model has at least as many modes as there are minor-joints in the arm. Another major} challenge is related to \emph{posterior's parameterization complexity} {$M(T):=|\mathcal{D}_k|$, which we subsequently call the dictionary size} (step 10 in Algorithm \ref{alg:PSRL}) which is used to parameterize the posterior distribution. {We note that $M(T)=\Omega(T)$ for the PSRL \cite{osband2014model} and MPC-PSRL \cite{pmlr-v139-fan21b} algorithms which are state of the art approaches. } 

 To alleviate the Gaussian restriction, we consider an alternative metric of evaluating the distributional estimation error, namely, the { kerneralized Stein discrepancy (KSD). Additionally, that KSD is easy to evaluate under appropriate conditions on the target distribution, i.e., {the target distribution is smooth}, one can compare its relative quality as a function of which data is included in the posterior. Doing so allows us to judiciously choose which points to retain during the learning process in order to ensure small Bayesian regret.} To our knowledge, this work is the first to deal with the compression of posterior estimate in model-based RL settings along with provable guarantees. These aspects are derived in detail in the following section. 

\section{Proposed Approach}
\subsection{Posterior Coreset Construction via KSD}
The core of our algorithmic development is based upon the computation of Stein kernels and KSD to evaluate the merit of a given transition model estimate, and determine which past samples to retain. Doing so is based upon the consideration of integral probability metrics (IPM) rather than total variation (TV) distance. This allows us to employ Stein's identity, which under a hypothesis that the score function of the target (which is gradient of log likelihood of target) is computable. This approach is well-known to yield methods to improve the sample complexity of Markov Chain Monte Carlo (MCMC) methods \cite{stein_point_Markov}. That this turns out to be the case in model-based RL as well is a testament to its power \cite{stein1956inadmissibility,ross2011fundamentals}. This method to approximate a target density $P$ consists of defining an IPM \cite{sriperumbudur2012empirical} based on a set $\mathcal{G}$ consisting of test functions on $\mathcal{X}\subset \mathbb{R}^{2 d_s + d_a}$, and is defined as:
\begin{equation}\label{eq:IPMs}
D_{\mathcal{G},P}(\{x_i\}_{i=1}^n)
\; := \;
\textstyle \sup_{g \in \mathcal{G}} \left| \frac{1}{n} \sum_{i=1}^n g(x_i) - \int_{\mathcal X} g \mathrm{d}P \right|, 
\end{equation}
where $n$ denotes the number of samples. 
%\red{sc : Should we describe the definition 7 by $\int_{\mathcal X} g \mathrm{d}Q$ and later adding one line that our empirical estimate $\frac{1}{n} \sum_{i=1}^n g(x_i)$}
We can recover many well-known probability metrics, such as total variation distance and the Wasserstein distance, through different choices of $\mathcal{G}$. Although IPMs efficiently quantify the discrepancy between an estimate and target, \eqref{eq:IPMs} requires $P$ to evaluate the integral, which may be unavailable.

To alleviate this issue, Stein's method restricts the class of test functions $g$ to be those such that $\mathbb{E}_P [g(z)] = 0$. In this case, IPM \eqref{eq:IPMs} only depends on the Dirac-delta measure $(\delta)$ from the stream of samples, removing dependency on the exact integration in terms of $P$. Then, we restrict the class of densities to those that satisfy this identity, which are simply those for which we can evaluate the score function of the target distribution. {Surprisingly, in practice, we need not evaluate the score function of the true posterior, but instead only the score function of estimated posterior, for this approach to operate \cite[proposition]{liu2016kernelized}.  Then, by supposing that the true density is smooth, the IPM can be evaluated in closed form through the kernelized Stein discrepancy (KSD) as a function of the \emph{Stein kernel} \cite{liu2016kernelized}, to be defined next.}
%\red{sc : Stein’s method creates discrepancy measures between two distributions that require only the unnormalized density of one and samples from the other (ref : https://arxiv.org/pdf/1904.04478.pdf. Might be helpful.}
%
\begin{algorithm}[tb]
  \caption{{\bf K}ernelized {\bf S}tein Discrepancy-based Posterior Sampling for {\bf RL} (\algo)  }
  \label{alg:MPC-PSRL}
\begin{algorithmic}[1]
  \STATE \textbf{Input} : Episode length $H$, Total timesteps $T$, Dictionary $\mathcal{D}$, prior distribution $\phi=\{\mathcal{P},\mathcal{R}\}$ for true MDP $M^*$, planning horizon $\tau$ for MPC Controller, thinning budget $\{\epsilon_{k}\}_{k=1}^K$
  \STATE \textbf{Initialization} : Initialize dictionary  $\mathcal{D}_1$ at with random actions from the controller as $\mathcal{D}_{1} :=\{s_{1,1},a_{1,1},r_{1,1},\cdots,s_{1,H},a_{1,H},r_{1,H} \}$, posterior $\phi_{\mathcal{D}_{1}}=\{\mathcal{P}_{\mathcal{D}_{1}},\mathcal{R}_{\mathcal{D}_{1}}\}$
  \FOR{{Episodes} $k=1$ to $K$}
  %
%   $M^k\sim \phi(\cdot|\mathcal{D}_k)$
  \STATE \textbf{Sample} a transition $P^k \sim \mathcal{P}_{\mathcal{D}_{k}}$ and reward model $r^k \sim \mathcal{R}_{\mathcal{D}_{k}}$ and initialize empty $\mathcal{C}=[]$
  \FOR{{timesteps} $i=1$ to $H$}
  %
  %\STATE \textbf{Generate} \text{$N$ Sample action sequences} $a_{k,i:k,i+\tau} \sim CEM(.)$ from the MPC controller with planning horizon $\tau$ [\textcolor{red}{check}]
  %
  \STATE \textbf{Evaluate} optimal action sequence $a^{*}_{k,i:k,i+\tau} = \arg\max_{a_{k,i:k,i+\tau}} \sum_{t=i}^{i+\tau} \mathbb{E}[r(s_{k,t}, a_{k,t})]$
  \STATE \textbf{Execute} $a^{*}_{k,i}$ from the optimal sequence $a^{*}_{k,i:k,i+\tau}$
  \STATE \textbf{Update} $\mathcal{C} \leftarrow \mathcal{C} \cup \lbrace(s_{k,i}, a_{k,i}, s_{k,i+1}, r_{k,i})\rbrace$
  \ENDFOR
    \STATE \textbf{Update} dictionary $\widetilde{\mathcal{D}}_{k+1} \leftarrow {\mathcal{D}}_{k} \cup {\mathcal{C}}$
    \STATE \textbf{Perform} thinning operation  (cf. Algorithm \ref{alg: inner loop})
\begin{align}
    (\phi_{{\mathcal{D}}_{k+1}},{\mathcal{D}}_{k+1}) =  \text{KSD-Thinning}(\phi_{\widetilde{\mathcal{D}}_{k+1}},\widetilde{\mathcal{D}}_{k+1},\epsilon_k) \nonumber
\end{align}

  \ENDFOR
\end{algorithmic}
\end{algorithm}

To be more precise, we define each particle $h_i$ as a tuple of the form $h_i:=(s_i,a_i,s'_i) \in \mathbb{R}^d$ (with $d=2d_s + d_a$) and the state-action tuple $\hat{h}_i:=(s_i,a_i) \in R^{d_s + d_a}$ . We would like a transition model estimate over past samples, and some appropriately defined test function $g$. This test function turns out to be a Stein kernel $\kappa_0$, which is explicitly defined in terms of base kernel $\kappa$, e.g., a Gaussian, or inverse multi-quadratic associated with the RKHS that imposes smoothness properties on the estimated transition model \cite{liu2016kernelized}. 
The explicit form of the Stein kernel $\kappa_0$ is given as follows
\begin{align}
    \label{eq: stein kernel}
    \kappa_0({h_i},\!{h_j}) = & s_P({h_i})^Ts_P({h_j})\kappa({h_i},\!{h_i}) \!\!+\!\!s_P({h_j})^T\nabla_{h_i} \kappa({h_i},\!{h_j})\nonumber
    \\
        &\hspace{-1cm} +s_P({h_i})^T\nabla_{h_j} \kappa(h_i,h_j)
         +\sum_{l=1}^d \frac{\partial^2 \kappa({h_i},{h_j})}{\partial {h_j}(l)\partial{h_j}(l)}\; ,
\end{align}
where $h_j(l)$ is the $l^{th}$ element of the $d-$dimensional vector and $s_{P}(h_i):=\nabla_{h_i}\log P(h_i)$ is the score function of true transition model $P$. Observe that this is only evaluated over \emph{samples}, and hence the score function of the true transition model is unknown. 
The key technical upshot of employing Stein's method is that we can now evaluate the integral probability metric of posterior $\phi_{\mathcal{D}_k}:=\phi(\cdot|\mathcal{D}_k)$ parameterized by dictionary $\mathcal{D}_k$ through the KSD, which is efficiently computable:
\begin{align}\label{KSD_estimate}
    \text{KSD}(\phi_{\mathcal{D}_k}):= \sqrt{\frac{1}{|\mathcal{D}_k|^2}\sum_{h_i,h_j}\kappa_0(h_i,h_j)} \; .
\end{align}
Therefore, we no longer require access to the true unknown target transition model of the MDP in order to determine the quality of a given posterior estimate of unknown target $P$. {This is a major merit of utilizing Stein's method in MBRL}, and allows us to improve the regret of model-based RL methods based on posterior sampling. 

This the previous point is distinct from the computational burden of storing dictionary $\mathcal{D}_k$ that parameterizes $\phi_{\mathcal{D}_k}$ and evaluating the optimal value function according to the current belief model \eqref{optimal_policy}. After this novel change (see Lemma \ref{lemma_1} in Sec. \ref{regret_analysis}), we can utilize the machinery of KSD to derive the regret rate for the proposed algorithm in this work  (cf. Algorithm \ref{alg:MPC-PSRL}) in lieu of concentration inequalities, as in \cite{pmlr-v139-fan21b}. We shift to the computational storage requirements of the posterior in continuous space next.

% \textcolor{blue}{To be specific, it allows to replace the distance to target distribution in the right hand side of \cite[Lemma 2]{pmlr-v139-fan21b} with a KSD term which can be evaluated in closed form via \eqref{KSD_estimate}. After this novel change (see Lemma \ref{lemma_1} in Sec. \ref{regret_analysis}), we can utilize the machinery of KSD posterior consistency to derive the regret rate for the proposed algorithm in this work  (cf. Algorithm \ref{alg:MPC-PSRL}) rather than using concentration inequalities as in \cite{pmlr-v139-fan21b}. }

\textbf{KSD Thinning:} We develop a principled way to avoid the requirement that the dictionary $\mathcal{D}_k$ retains all information from past episodes, and is instead parameterized by a coreset of statistically significant samples. More specifically, observe that in step $10$ and $11$ in PSRL (see Algorithm \ref{alg:PSRL} in Appendix \ref{sec:Prelim}), the dictionary at each episode $k$ retains $H$ additional points, i.e.,  $|\mathcal{D}_{k+1}|=|\mathcal{D}_{k}|+H$.  Hence, as the number of episodes experienced becomes large, the posterior representational complexity grows linearly and unbounded with episode index $k$. On top of that, the posterior update in step 11 in PSRL (cf. Algorithm \ref{alg:PSRL}) is also parameterized by {data} collected in $\mathcal{D}_{k+1}$. For instance, if the prior is assumed to be Gaussian, the posterior update of step 11 in PSRL (cf. Algorithm \ref{alg:PSRL}) boils down to GP posterior parameter evaluations which is of complexity $\mathcal{O}(|\mathcal{D}_{k}|^3)$ for each $k$ \cite{rasmussen2004gaussian}. 

To deal with this bottleneck, we propose to sequentially remove those particles from $\mathcal{D}_{k+1}$ that contribute least in terms of  KSD. This may be interpreted as projecting posterior estimates onto ``subspaces" spanned by only statistically representative past state-action-state triples. This notion of representing a nonparametric posterior using only most representative samples has been shown to exhibit theoretical and numerical advantages in probability density estimation  \cite{campbell2018bayesian,campbell2019automated}, Gaussian Processes \cite{koppel2021consistent}, and Monte Carlo methods \cite{elvira2016adapting}. Here we introduce it for the first time in model-based RL, which allows us to control the growth of the posterior complexity, which in turn permits us to obtain computationally efficient updates. 
\begin{algorithm}[t]
\caption{Posterior Coreset with {\bf KSD Thinning} for \textbf{R}einforcement {\bf L}earning (KSD-Thinning)}
\begin{algorithmic}[1]
\label{alg: inner loop}
\STATE \textbf{Input:} $(q_{\mathcal{W}},\mathcal{W},\epsilon)$
\STATE \textbf{Require:} Target score function  %\textcolor{red}{is it true target or just estimate, need to highlight that}
\STATE \textbf{Compute} the reference KSD as  $\alpha:=\text{KSD}(q_{\mathcal{W}})$ via \eqref{KSD_estimate}
\WHILE{$\text{KSD}(q_{\mathcal{W}})^2<\alpha^2+\epsilon$} % and $|\dict|>S$}
\STATE \textbf{Compute} the least influential point ${x}_j$ as the minimial $h_i\in\widetilde{\mathcal{D}}_{k+1}$ \eqref{eq: ksd goal}% [\textcolor{red}{this is not clear}] 
\IF{$\text{KSD}(q_{\mathcal{W} \setminus \{{x}_j\}})^2<\alpha^2+\epsilon$}
\STATE Remove the least influential point, set $\mathcal{W} = \mathcal{W}\setminus\{{x}_j\}$
\ELSE
\STATE Break loop
\ENDIF
\ENDWHILE \\
\STATE \textbf{Output} thinned dictionary $\mathcal{W}$ satisfying $\text{KSD}(q_{\mathcal{W}})^2<\alpha^2+\epsilon$
\end{algorithmic}
\end{algorithm}

To be more specific, suppose we are at episode $k$ with dictionary $\mathcal{D}_k$ associated with posterior  $\phi_{\mathcal{D}_k}$, and we denote the dictionary after update as $\widetilde{\mathcal{D}}_{k+1}=\mathcal{D}_{k}+H$ and corresponding posterior as ${\phi}_{\widetilde{\mathcal{D}}_{k+1}}$. For a given dictionary $\mathcal{D}_k$, we can calculate the KSD of posterior $\phi_{\mathcal{D}_k}$ to target via \eqref{KSD_estimate}. 
%
%\begin{align}\label{KSD_estimate}
%    \text{KSD}(\phi_{\mathcal{D}_k}):= \sqrt{\frac{1}{|\mathcal{D}_k|^2}\sum_{h_i,h_j}\kappa_0(h_i,h_j)},
%\end{align}
%
We note that \eqref{KSD_estimate} goes to zero as $k\rightarrow \infty$ due to the posterior consistency conditions \cite{gorham2015measuring}. At each episode $k$, after performing the dictionary update to obtain $\widetilde{\mathcal{D}}_{k+1}$ (step 10 in Algorithm \ref{alg:MPC-PSRL}), we propose to thin dictionary $\widetilde{\mathcal{D}}_{k+1}$ such that
\begin{align}    \label{eq: ksd goal}
    \text{KSD}(\phi_{\mathcal{D}_{k+1}})^2< \text{KSD}(\phi_{\widetilde{\mathcal{D}}_{k+1}})^2+\epsilon_{k+1},
\end{align}
where $\mathcal{D}_{k+1}$ is the dictionary following thinning and $\epsilon_k>0$ is a scalar parameter we call the thinning budget proposed. This means the posterior defined by compressed dictionary $\phi_{\widetilde{\mathcal{D}}_{k+1}}$ is at most $\epsilon_{k+1}$ in KSD from its uncompressed counterpart. See \cite{hawkins2022online} for related development of this compression routine in the context of MCMC.  We will see in the regret analysis section (cf. Sec. \ref{regret_analysis}) how $\epsilon_{k}$ permits us to trade off regret and dictionary size in practice. \eqref{eq: ksd goal} may be succinctly stated as
\begin{align}
    (\phi_{{\mathcal{D}}_{k+1}},{\mathcal{D}}_{k+1}) =  \text{KSD-Thinning}(\phi_{\widetilde{\mathcal{D}}_{k+1}},\widetilde{\mathcal{D}}_{k+1},\epsilon_k).
\end{align}
We summarize the proposed algorithm in Algorithm \ref{alg:MPC-PSRL} with compression subroutine in Algorithm \ref{alg: inner loop}, where {\algo} is an abbreviation for Kernelized Stein Discrepancy Thinning for Model-Based Reinforcement Learning. Please refer to the discussion in Appendix \ref{sec:Prelim} for MPC-based action selection. 
\section{Bayesian Regret Analysis} \label{regret_analysis}
In this section, we establish the regret of Algorithm \ref{alg:MPC-PSRL}. Our analysis builds upon \cite{osband2014model}, but exhibits fundamental departures in the sense that we consider an alternative metric for quantifying the posterior estimation error using IPMs that exploit's salient structural properties of Stein's method, which additionally provides a basis for establishing tradeoffs between regret and posterior representational complexity which is novel in this work. Begin then by restating the model error at episode $k$ from \eqref{regret_0} as \cite{osband2014model}
\begin{align}
    {\Delta}_k = & \int \rho(s_1)( {V_{{\mu}^*}^{{{M}^*}}(s_1)-V_{\tilde{\mu}^k}^{{M^*}}(s_1))} ds_1  \nonumber\\
      = & \underbrace{\int \rho(s_1)( V_{{\mu}^*}^{{{M}^*}}(s_1) -V_{{\mu}^k}^{{{M}}^k}(s_1)) ds_1}_{=:\Delta^I_{k}} 
      \nonumber \\
      & \quad \quad \quad \quad \quad + \underbrace{\int  (V_{{\mu}^k}^{{{M}}^k}(s_1)  -  V_{{\mu}^k}^{{M^*}}(s_1)   )ds_1}_{=:\Delta^{II}_{k}},
\end{align}
 where we add and subtract the term $\int \rho(s_1)  (V_{{\mu}^k}^{{{M}}^k}(s_1))ds_1$,  ${\mu}^k$ represents the optimal policy (cf. \eqref{optimal_policy}) under constructed model ${{M}}^k$ from the thinned posterior we obtain via procedure proposed in Algorithm \ref{alg:MPC-PSRL}. Hence, the regret for episode $k$ can be decomposed as $ {\Delta}_k= \Delta^I_{k}+\Delta^{II}_{k}$ which implies 
 \begin{equation}\label{regret}
{Regret}_{T}=\sum_{k=1}^{\lceil\frac{T}{H}\rceil} \Delta^I_{k}+\sum_{k=1}^{\lceil\frac{T}{H}\rceil} \Delta^{II}_{k},
\end{equation}
where $\lceil\frac{T}{H}\rceil$ denotes the number of total episodes $T$ is the total number of timesteps and $H$ is the number of timesteps per episode.  The equation in \eqref{regret} matches with the regret decomposition in \cite{osband2014model} but there is a fundamental departure: specifically,  \eqref{regret} the sample $M_k$ for episode $k$ is sampled from the \textbf{\emph{thinned}} posterior following the procedure proposed in Algorithm \ref{alg:MPC-PSRL}. Similar to prior works \cite{osband2014model,pmlr-v139-fan21b}, since we are also performing posterior sampling $M^k\sim \phi(\cdot~|~ \mathcal{D}_k)$ for each $k$, hence we note that $\mathbb{E}[\Delta_k^I]=0$. 
Next, we take the expectation on both sides in \eqref{regret} to write
 \begin{equation}\label{regret_1}
\mathbb{E}[{Regret}_{T}]=\sum_{k=1}^{\lceil\frac{T}{H}\rceil} \mathbb{E}[\Delta^{II}_{k}],
\end{equation}
which implies that the first term in \eqref{regret} is null, which allows us to shift focus to analyzing the expected value of $\Delta^{II}_{k}$ for each $k$. To proceed with the analysis, we relate the estimation error of the future value function to the KSD in Lemma \ref{lemma_1}, which is a key novelty of this work that exploit's Stein's method. { To keep the exposition simple, We first derive Lemma \ref{lemma_1} as follows. }
%for $P^k \sim \mathcal{P}_{\mathcal{D}_{k}}$ and we have  $\phi_{\mathcal{D}_{k}}=\{{P}_{\mathcal{D}_{k}},{R}_{\mathcal{D}_{k}}\}$ (cf. Algorithm \ref{alg:MPC-PSRL}). Next we state Lemma \ref{lemma_1}}.

\begin{lemma}(Lipschitz in Kernel Stien Discrepancy)\label{lemma_1}
Recall the definition of the future value function in \eqref{value_function}. {Under the assumption that posterior distributions are continuously differentiable (also called smooth)} 
the future value function estimation error of $P^k$ with respect to $P^*$ is upper-bounded in terms of the KSD of $P^k$
\begin{align}\label{lips_KSD}
\! \! \!\!\!\! U_{i}^k({P}^k(\hat{h}_i))-U_{i}^k(P^*(\hat{h}_i)) 
&\leq  HR_{\text{max}} {\text{KSD}(P^k(h_i))},
\end{align}
 for all  $i$ and  $k$.
\end{lemma}
See Appendix \ref{proof_lemma_1} for proof. Observe that Lemma \ref{lemma_1} is unique to this work, and is a key point of departure from  \cite{osband2014model,pmlr-v139-fan21b}. An analogous inequality in  \cite{osband2014model} mandates that the future value function is  Lipschitz continuous without an explicit value of Lipschitz parameter. This implicit dependence is made explicit in \cite[Lemma 2]{pmlr-v139-fan21b}, where the Lipschitz parameter is explicitly shown to depend on episode length $H$ under the assumption that underlying model in Gaussian. Furthermore, an important point to note that both in \cite{osband2014model,pmlr-v139-fan21b}, this upper bound on the future value function is provided in terms of total variation norm between the distributions $P^k$ and $P^*$. In contrast, we take a different route based on Stein's method that alleviates the need for Gaussian assumption for the underlying transition model, and replaces the TV norm with KSD  of $P^k(h_i)$. 

{Observe that the right hand side of \eqref{lips_KSD} depends on the KSD of joint posterior $P^k(h_i)$ which we define for $h_i$. The Stein's method allows us to consider the KSD of joint because the score function of conditional distribution and joint distribution are the same and does not depend upon the normalizing constant.}  The salient point to note here is that we do not require access to $P^*$ to evaluate the right hand side of   \eqref{lips_KSD}. True to our knowledge, this is the first time that power of Stein's methods is being utilized in model-based RL. Next, we proceed towards deriving the regret for Algorithm \ref{alg:MPC-PSRL}. First, we need an additional result which upper bounds the KSD of current posterior $\text{KSD}(\phi_{\mathcal{D}_k})$  at episode $k$ which we state next in Lemma \ref{lemma_2}. %

\begin{lemma}(KSD Upper bound)  \label{lemma_2} Under the Assumptions of Lemma \ref{lemma_1} and thinning budget $\epsilon_{k}=\frac{\log(k)}{f(k)^2}$,  for the iterates of proposed Algorithm \ref{alg:MPC-PSRL}, it holds that 
\begin{align}
\mathbb{E}\left[\text{KSD}(\Lambda_{\mathcal{D}_k})\right] 
= & \mathcal{O} \left(\frac{\sqrt{k\log(k)}}{f(k)}\right),
\end{align}
where $f(k)$ lower-bounds the growth rate of the {posterior’s parameterization complexity (coreset size)} as $|\mathcal{D}_k|\geq f(k)$.  
\end{lemma}
See Appendix \ref{proof_lemma_2}  for proof. The inequality established in Lemma \ref{lemma_2} relates the KSD to the number of episodes experienced by a model-based RL method, and may be interpreted as an adaption of related rates of posterior contraction in terms of KSD that have appeared in MCMC \cite{stein_point_Markov}. In particular, with this expression, we can relate the expected value of the KSD of current thinned posterior $\phi_{\mathcal{D}_k}$ to the target density for each episode $k$. For the statistical consistency of the posterior estimate, we note that it is sufficient to show that $\mathbb{E}\left[\text{KSD}(\phi_{\mathcal{D}_k})\right] \rightarrow 0$ as $k\rightarrow \infty$, which imposes a lower bound on the dictionary size growth rate $f(k)>\sqrt{k\log(k)}$ from the statement of Lemma \ref{lemma_1} required for convergence. This result communicates that it is possible to achieve statistical consistency without having a linear growth in the dictionary size that is a drawback of prior art \cite{osband2014model,pmlr-v139-fan21b}. Next, we ready to combine Lemmas \ref{lemma_1} - \ref{lemma_2} to establish our main result.

\begin{theorem}[Regret and Coreset Size Tradeoff {\algo}]\label{thm: decaying budget}
Under the thinning budget $\epsilon_{k}=\frac{\log(k)}{f(k)^2}$ and {coreset} size growth condition $f(k) = \sqrt{k^{\alpha+1} \log(k)}$ where $\alpha\in(0,1]$, the total Bayes regret for our KSD based posterior thinning algorithm for model-based RL (cf. Algorithm \ref{alg:MPC-PSRL}) is given by
\begin{align}
\label{eq: decaying budget conclusion}
\mathbb{E}[\text{Regret}_{T}] = \mathcal{O}\left({d} T^{1-\frac{\alpha}{2}} H^{1+\frac{\alpha}{2}}\right)
\end{align}
and {coreset size} ($M(T)$) order is given by
%
%\begin{align}
  $ M(T) = \tilde\Omega\left(\sqrt{T^{1+\alpha}}\right)$,
%\end{align}
%
where $T$ denotes the total number of state-action pairs processed, and $H$ is the length of each episode.  

\end{theorem}

\begin{figure*}[ht]
     \centering
     \begin{subfigure}[b]{0.28\textwidth}
         \centering
\includegraphics[width=\textwidth]{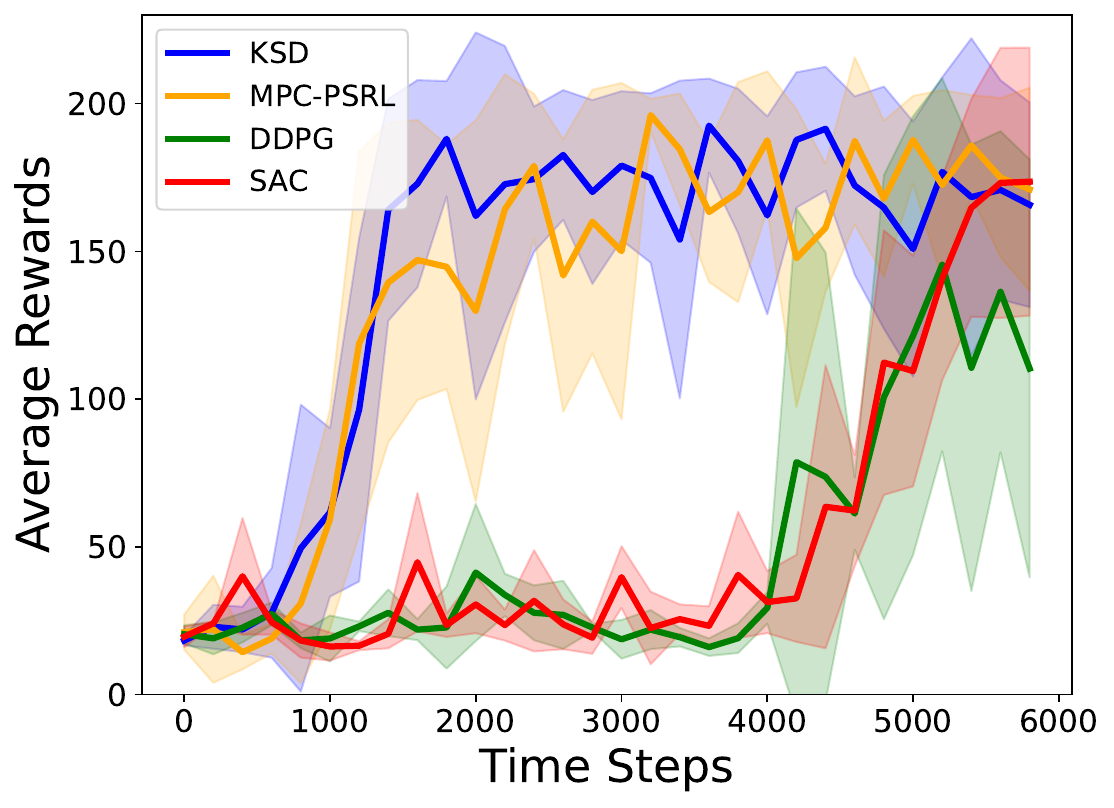}
         \caption{}
         \label{fig:five over x1}
     \end{subfigure}
          \hfill
     \begin{subfigure}[b]{0.28\textwidth}
         \centering
\includegraphics[width=\textwidth]{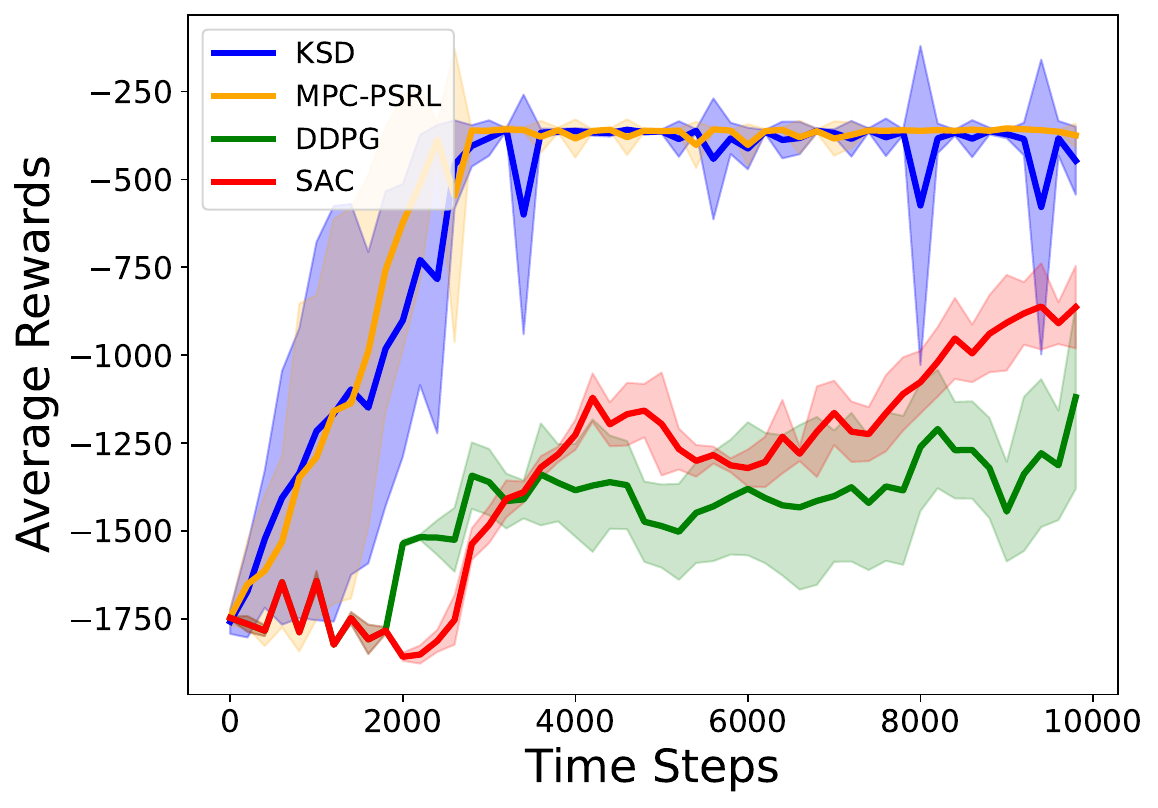}
         \caption{}
         \label{fig:five over x2}
     \end{subfigure}
     \hfill
          \begin{subfigure}[b]{0.28\textwidth}
         \centering
\includegraphics[width=\textwidth]{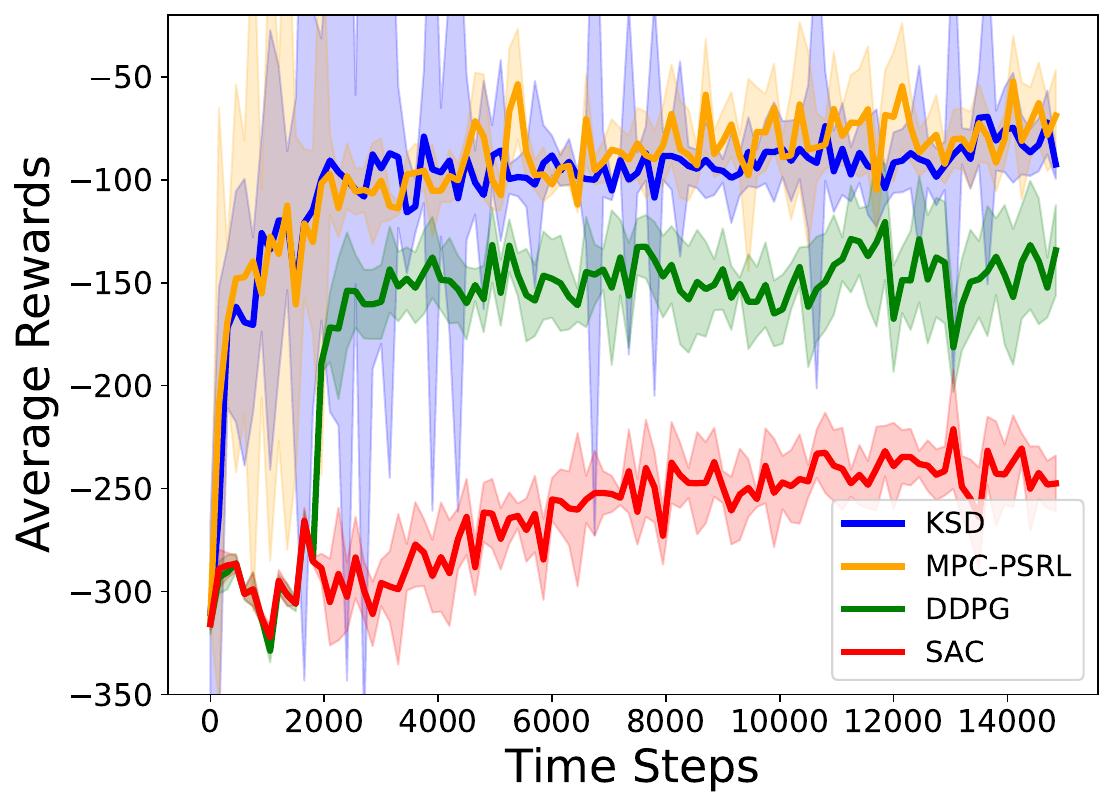}
         \caption{}
         \label{fig:five over x3}
     \end{subfigure}\\
          \hfill
     \begin{subfigure}[b]{0.28\textwidth}
         \centering
\includegraphics[width=\textwidth]{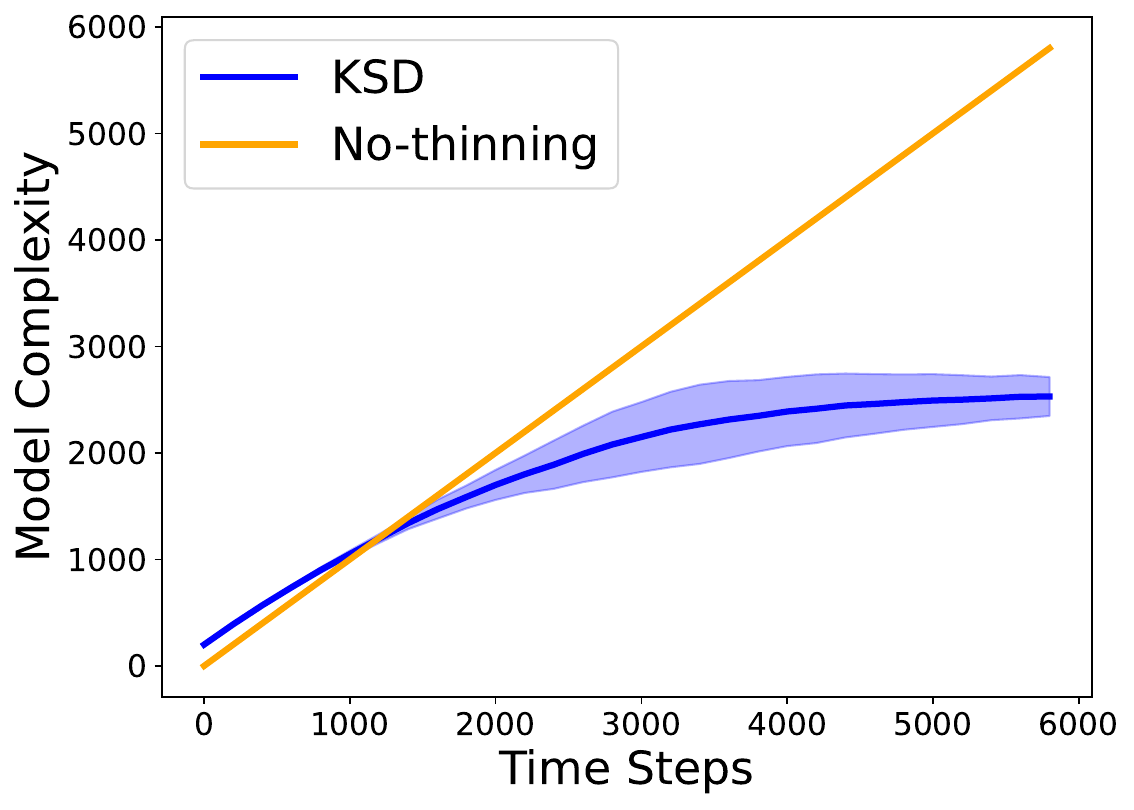}
         \caption{}
         \label{fig:five over x4}
     \end{subfigure}
          \hfill
          \begin{subfigure}[b]{0.28\textwidth}
         \centering
\includegraphics[width=\textwidth]{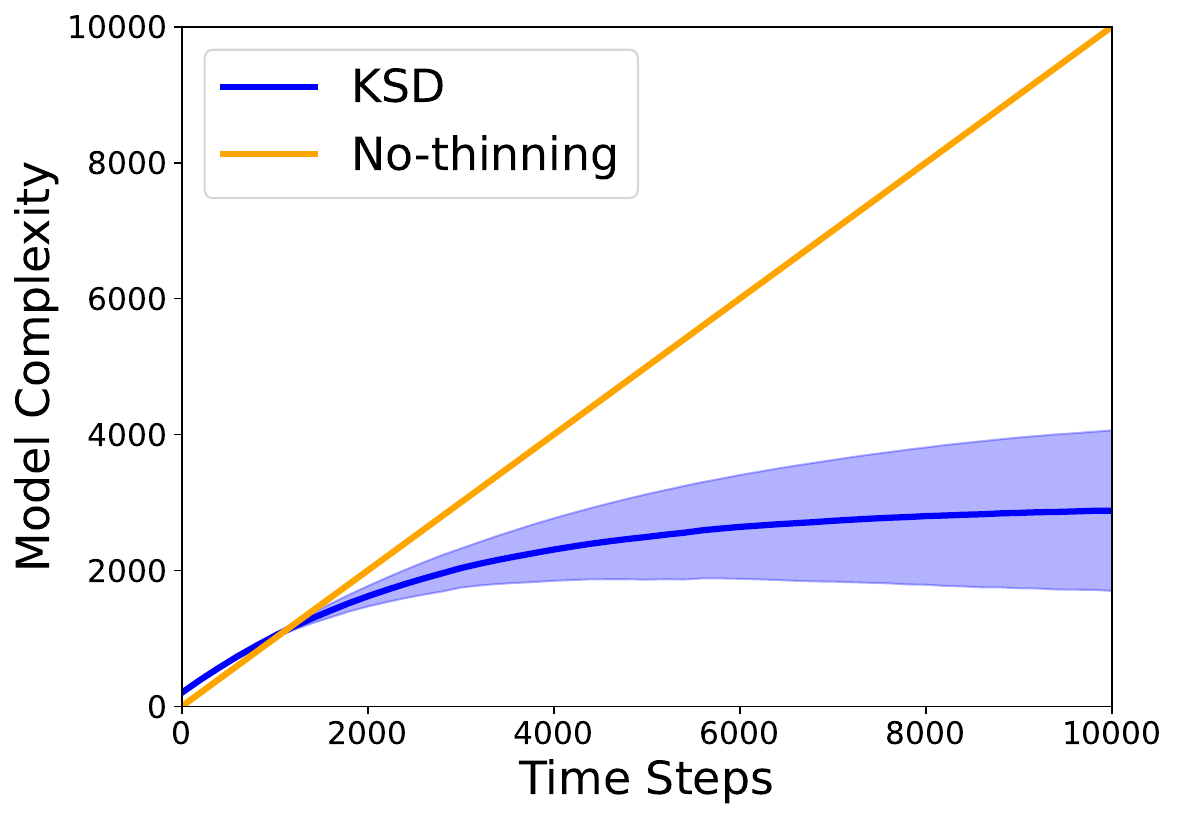}
         \caption{}
         \label{fig:five over x5}
     \end{subfigure}
          \hfill
     \begin{subfigure}[b]{0.28\textwidth}
         \centering
\includegraphics[width=\textwidth]{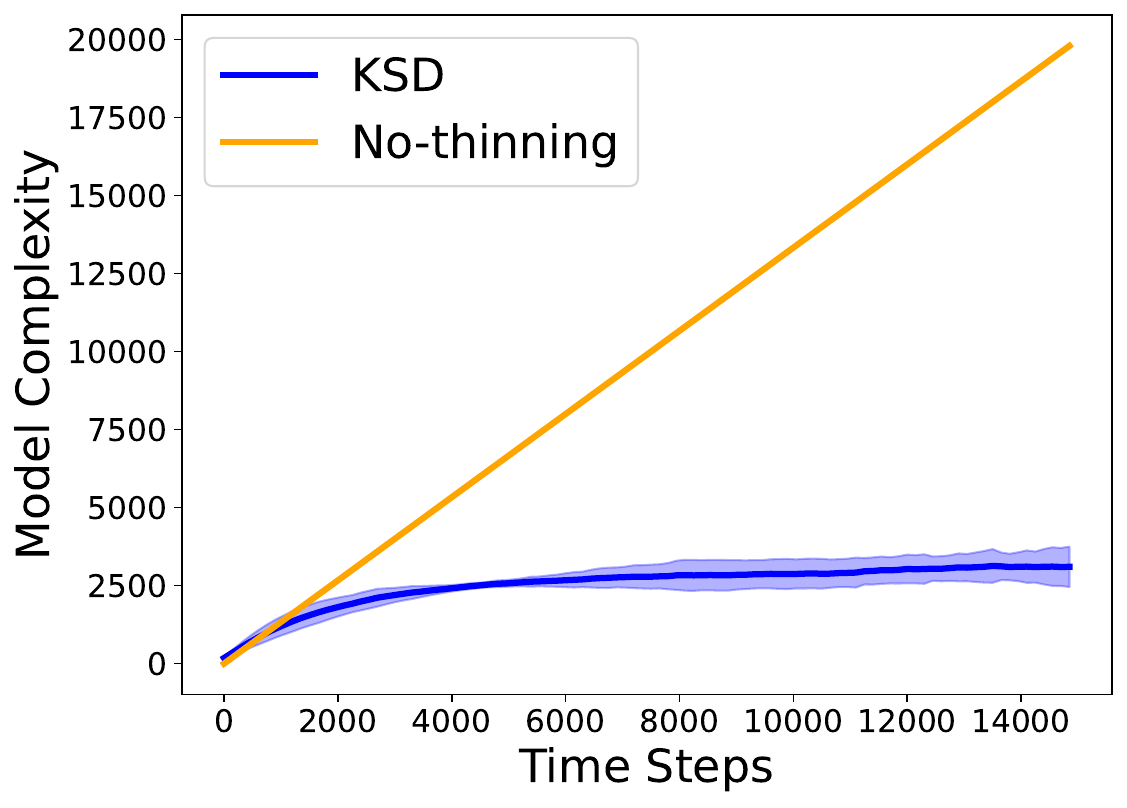}
         \caption{}
         \label{fig:five over x6}
     \end{subfigure}
        \caption{{(a)-(c)} compares the average cumulative reward return achieved by the proposed {\algo} (shown in blue) algorithm with MPC-PSRL \cite{pmlr-v139-fan21b}, SAC \cite{haarnoja2018soft}, and DDPG \cite{barth2018distributed} for modified {Cartpole}, {Pendulum}, and {Pusher} without rewards. Figures with rewards are shown in the Appendix \ref{experimental_results}.
        {(d)-(f)} compares the model-complexity.  We note that {\algo} is able to achieve the maximum average reward at-par with the current SOTA MPC-PSRL with drastically reduced model complexity. Solid curves represent the average across five trials (seeds), shaded areas correspond to the standard deviation amongst the trials.}
        \label{fig:threegraphs_main_body}
\end{figure*}
The proof is provided in Appendix \ref{proof_theorem_1}. To establish Theorem \ref{thm: decaying budget}, we start with the upper bound on the term $\Delta_k^{II}$ via utilizing the result from the statement of Lemma \ref{lemma_1}. Then we upper bound the right hand side of \eqref{lips_KSD} via  the relationship between KSD and the number of past samples (Lemma \ref{lemma_2}). 

{\textbf{Remark 1 (Dependence on Dimension):} Prior results \cite{osband2014model,pmlr-v139-fan21b} exhibit a linear dependence on the input dimension $d$, which matches our dependence. However, these results require the posterior to belong to a symmetric class of distributions. We relax this assumption and only require our posterior to be smooth, a substantial relaxation. }

\textbf{Remark 2 (Tradeoff between regret and model complexity):} An additional salient attribute of Theorem \ref{thm: decaying budget} is the introduction of compression budget $\epsilon_k$ which we specify in terms of tunable parameter $\alpha$. This quantity determines the number of elements that comprise the thinned posterior during model-based RL training. We provide a more quantitative treatment in the Table \ref{tradeoff}.
\begin{table}[h]
\centering
\resizebox{0.5\columnwidth}{!}{\begin{tabular}{|c | c | c|} 
 \hline
 $\alpha$ & $\mathbb{E}[\text{Regret}_{T}]$ & $M(T)$\\ [0.5ex] 
 \hline\hline
0 & $\mathcal{O}(dHT)$   &  $\widetilde{\Omega}(\sqrt{T})$\\ 
 \hline
 0.5 & $\mathcal{O}(dH^{\frac{7}{4}} T^{\frac{1}{4}})$  & $\widetilde \Omega({ T^{3/4}})$ \\
 \hline
 1 & $\mathcal{O}(dH^{\frac{3}{2}} T^{\frac{1}{2}})$ & $ \widetilde\Omega( T)$ \\
 \hline
\end{tabular}}
\caption{Tradeoff for different values of $\alpha$.}
\label{tradeoff}
%\vspace{-5mm}
\end{table}
Here, in the table, we present the regret analysis for our Efficient Stein-based PSRL algorithm. Observe that we match the best known prior results of PSRL with $\alpha =1$ as shown in \cite{pmlr-v139-fan21b} in-terms of $O (d H^{\frac{3}{2}} T^{\frac{1}{2}})$, but with relaxed conditions on the posterior, allowing the approach to apply to a much broader class of problems. Moreover, for $\alpha<1$, we obtain a superior tradeoff in model complexity and regret. 
%However, this dependence on the horizon is achieved \emph{without} any dependence on $d$. 
%Hence, we improve the best-known regret bounds for efficient model-based posterior sample based reinforcement learning \cite{osband2014model,pmlr-v139-fan21b}. 
Now, from a statistical consistency perspective a posterior is consistent for $\beta$ if the posterior distribution on $\beta$ concentrates in neighborhoods of the true value. 

\section{Experiments}

In this section, we present a detailed experimental analysis of {\algo}  as compared to state of the art model-based and model-free RL methods on several continuous control tasks in terms of training rewards, model complexity and KSD convergence. First we discuss the different baseline algorithms to which we compare the proposed {\algo}. Secondly, we details the experimental environments, and then we empirically validate and analyze the performance of {\algo} in detail. 
%convergence of the Kernel Stein discrepancy (squared) over the iterations for our KSD-SEPSRL algorithm. 

%\subsection{}\label{baselines}

\textbf{Baselines.} For comparison to other model free approaches, we compare against MPC-PSRL method propose in \cite{pmlr-v139-fan21b}. There are other popular model based methods in literature such as MBPO \cite{mbpo} and PETS \cite{chua2018deep} but MPC-PSRL is already shown to outperform them in \cite[Fig. 1]{pmlr-v139-fan21b}. Since the underlying environments are same, we just compare to MPC-PSRL and show improvements. 
For comparison to model-free approaches, we compare with Soft Actor-Critic (SAC) from \cite{haarnoja2018soft} 
and Deep Deterministic Policy Gradient (DDPG) \cite{barth2018distributed}.  

\textbf{Environment Details.} We consider continuous control environments Stochastic Pendulum, Continuous Cartpole, Reacher and Pusher with and without rewards of modified OpenAI Gym \cite{openai} \& MuJoCo environments \cite{Mujoco}. These environments are of different complexity and provide a good range of performance comparisons in practice. See Appendix \ref{experimental_results} for additional specific details of the environments and architecture.

\textbf{Discussion.} Fig. \ref{fig:threegraphs_main_body} compares the average reward return (top row) and model complexity (bottom row) for Cartpole, Pendulum, and Pusher, respectively. We note that {\algo} performs equally good or even better as compared to the state-of-the-art MPC-PSRL algorithm with a significant reduction in  model complexity (bottom row in Fig. \ref{fig:threegraphs_main_body}) consistently across different environments. From the model complexity plots, we remark that {\algo} is capable of automatically selecting the data points and control the dictionary growth across different environments which helps to achieve same performance in terms of average reward with fewer dictionary points to parameterize posterior distributions. This also helps in achieving faster compute time in practice to perform the same task as detailed in the  Appendix \ref{experimental_results}. We also show improvements of our algorithm over MPC with fixed buffer size in Appendix \ref{experimental_results} with both random and sequential removal.

% \begin{figure}[h]
% \includegraphics[width=4.8cm]{reacher_nrewards_nips.pdf}
% \caption{Comparison of our \algo  with MPC-PSRL restricted with fixed size buffer on Reacher which clearly shows improvement of our \algo}
% \label{fig:improve_reacher}
% \end{figure}

\section{Conclusions}

In this work, we develop a novel Bayesian regret analysis for model-based RL that can scale efficiently in continuous space under more general distributional settings and we achieve this objective by constructing a coreset of points that contributes the most to the posterior as quantified by KSD. Theoretical and experimental analysis bore out the practical utility of this methodology. 
%\clearpage
%\bibliographystyle{aaai23.sty}
\bibliography{aaai23}

%%%%%%%%%%%%%%%%%%%%%%%%%%%%%%%%%%%%%%%%%%%%%%%%%%%%%%%%%%%
\clearpage
\onecolumn
\appendix
\fontsize{11}{12}\selectfont

\section*{ \centering Appendix }

\section{Background}\label{sec:Prelim}
Let us write down the posterior sampling based reinforcement learning (PSRL) algorithm here in detail which is the basic building block of the research work in this paper \cite{osband2014model}. 
\begin{algorithm}[h]
  \caption{Posterior Sampling for Reinforcement Learning (PSRL) \cite{osband2014model} }
  \label{alg:PSRL}
\begin{algorithmic}[1]
  \STATE \textbf{Input} : Episode length $H$, Total timesteps $T$, Dictionary $\mathcal{D}_1$, prior distribution $\phi$ for $M^*$, i=1 
  \FOR{{episodes} $k=1$ to $K$}
  \STATE Sample $M^k\sim \phi(\cdot|\mathcal{D}_k)$ %
  \STATE Evaluate $\mu^{k}$ under $M^k$ via \eqref{optimal_policy} and initialize empty $\mathcal{C}=[]$
  \FOR{{timesteps} $i=1$ to $H$}
  \STATE Take action $a_i\sim\mu^{k}(s_i)$ 
    \STATE Observe $r_i$ and $s_{i+1}$ action $a_i\sim\mu^{k}(s_i)$ 
  \STATE Update $\mathcal{C} = \mathcal{C} \cup \lbrace (s_i, a_i, r_i, s_{i+1}) \rbrace$
  \ENDFOR
  \STATE  Update $\mathcal{D}_{k+1} = \mathcal{D}_{k} \cup \mathcal{C}$
    \STATE  Update posterior to obtain $\phi(\cdot~|~\mathcal{D}_{k+1}) $
    \ENDFOR
\end{algorithmic}
\end{algorithm}

\textbf{Model Predictive Control} : Model based RL planning with Model-predictive control (MPC) has achieved great success\cite{mpc} in several continuous control problems especially due to its ability to efficiently incorporate uncertainty into the planning mechanism. MPC has been an extremely useful mechanism for solving multivariate control problems with constraints \cite{camacho2013model} where it solves a finite horizon optimal control problem in a receding horizon fashion. The integration of MPC in the Model-based RL is primarily motivated due to its implementation simplicity, where once the model is learnt, the subsequent optimization for a sequence of actions is done through MPC. At each timepoint, the MPC applies the first action from the optimal action sequence under the estimated dynamics and reward function by solving $\arg\max_{a_{i:i+\tau}} \sum_{t=i}^{i+\tau} \mathbb{E}[r(s_{t}, a_{t})]$. However, computing the exact $\arg\max$ is non-trivial for non-convex problems and hence approximate methods like random sampling shooting \cite{nagabandi2018neural}, cross-entropy methods \cite{botev2013cross} are commonly used. For our specific case we use the Cross-entropy method for its effectiveness in sampling and data efficiency.

\section{Proof of Lemma \ref{lemma_1}}\label{proof_lemma_1}
\begin{proof}

Using the definition of the future value function in \eqref{value_function} with the Bellman Operator, we can write
\begin{align}
 U_{i}^k({P}^k(\hat{h}_i))-U_{i}^k(P^*(\hat{h}_i)) \leq & \max_{s}|V^{{M}_k}_{{\mu}_k}(s)|\cdot \|{P}^{k}(\cdot|\hat{h}_i)-P^*(\cdot|\hat{h}_i)\|
 \\
\leq & HR_{\text{max}}\|{P}^{k}(\cdot|\hat{h}_i)-P^*(\cdot|\hat{h}_i)\|, \label{here_norm}
\end{align}
, can we where we have utilized the absolute upper bound on the value function $|V(s)|\leq HR_{\text{max}}$. The norm in \eqref{here_norm} is the total variation  distance between the probability measures. In \cite[Lemma 2]{pmlr-v139-fan21b}, the right-hand side of \eqref{here_1} is further upper bounded by the distance between the mean functions after assuming that the underlying distributions are Gaussian. This is the point of departure in our analysis where we introduce the notion of Stein discrepancy to upper bound the total variation distance between the probability measures. 

First, we build the analysis for $d=1$ and later would extend it to multivariate scenarios. We start the analysis by showing that the total variation distance is upper bounded by the KSD for $d=1$. Let us define the notion of an integral probability metric between two distributions $p$ and $q$ as 
$$
D(p,q):=\sup_{f\in\mathcal{F}} \Bigg|\int fdp-\int fdq\Bigg|,
$$
where $\mathcal{F}$ is any function space. Now, if we restrict ourselves to a function space $\mathcal{F}':=\{f:\|f\|_{\infty}\leq 1\}$, then we boils down to the definition of total variation distance between $p$ and $q$ given by 
$$
TV(p,q):=\sup_{f\in\mathcal{F}'} \Bigg|\int fdp-\int fdq\Bigg|.
$$
We can restrict our function class to be $RKHS$ given by $\mathcal{H}$  and still write 
%which means $\mathcal{F}=\mathcal{H}$, then also I should be able to write the definition of TV norm between $p$ and $q$ such that
$$
TV(p,q):=\sup_{f\in\mathcal{H}'} \Bigg|\int fdp-\int fdq\Bigg|,
$$
where we define $\mathcal{H}':=\{f:\|f\|_{\infty}\leq 1\}$. We note that $\mathcal{H}'$ is uniquely determined by the kernel $\kappa(x,y)$ (e.g., RBF kernel) .
    Now, we know $\mathcal{H}'$ is a subset of RKHS $\mathcal{H}$. Now, we will try to upper bound the supremum over $\mathcal{H}'$ with supremum over a general class of functions which we call Stein class of functions as $\mathcal{H}_S$ \cite{liu2016kernelized}. Note that since $\mathcal{H}'$
$$
TV(p,q)=\sup_{f\in\mathcal{H}'} \Bigg|\int fdp-\int fdq\Bigg|\leq \sup_{f\in\mathcal{H}_S} \Bigg|\int fdp-\int fdq\Bigg|. 
$$
Now, for Stein class of functions $\mathcal{H}_S$, it holds that $\int fdp=0$. Therefore, we can write
$$
TV(p,q)=\sup_{f\in\mathcal{H}'} \Bigg|\int fdp-\int fdq\Bigg|\leq \sup_{f\in\mathcal{H}_S} \Bigg|\int fdp-\int fdq\Bigg|=\sup_{f\in\mathcal{H}_S} \Bigg|\int fdq\Bigg|\\
$$
Hence, we can write
$$
TV(p,q)\leq \sup_{f\in\mathcal{H}_S} \Bigg|\int fdq\Bigg|=: \text{KSD}(q).
$$
Utilizing the above upper bound into the right hand side of \eqref{here_norm}, we can write 
\begin{align}
 U_{i}^k({P}^k(\hat{h}_i))-U_{i}^k(P^*(\hat{h}_i)) \leq & HR_{\text{max}} \text{KSD}\left({P}^{k}(h_i)\right). \label{here_norm20}
\end{align}
The above result holds for $d=1$ case, and assuming that the variables are independent of each other in all the dimensions, we can naively write that in $d$ dimensions, we have 
\begin{align}
 U_{i}^k({P}^k(\hat{h}_i))-U_{i}^k(P^*(\hat{h}_i)) \leq & d HR_{\text{max}} \text{KSD}\left({P}^{k}(h_i)\right). \label{here_norm2}
\end{align}
Hence proved. 
\end{proof}

\section{Proof of Lemma \ref{lemma_2}}\label{proof_lemma_2}
Before starting the proof, here we discuss what is unique about it as compared to the existing literature. The kernel Stein discrepancy based compression exists in the literature \cite{hawkins2022online} but that is limited to the settings of estimating the distributions. This is the first time we are extending the analysis to model based RL settings. The samples in our setting are collected in the form of episodes following an optimal policy $\mu^k$ (cf. \eqref{optimal_policy}) for each episode. The analysis here follows a similar structure to the one performed in the proof of \cite[Theorem 1]{hawkins2022online} with careful adjustments to for the RL setting we are dealing with in this work. Let us now begin the proof. 
\begin{proof}
Since we are learning for both reward and transition model denoted by $\mathcal{R}$ and $\mathcal{P}$ and they both are parameterized by the same dictionary $\mathcal{D}_k$, we present the KSD analysis for $\Lambda_{\mathcal{D}_k}:=\{P^k, R^k\}$ without loss of generality. Further, for the proof in this section, we divide the $H$ samples collected in $k^{th}$  episode into $M>1$ number of batches, and then select the KSD optimal point  from each batch of $H/M$ samples similar to SPMCMC procedure. We will start with the one step transition at $k-1$ where we have the dictionary $\mathcal{D}_{k-1}$ and we update it to obtain $\widetilde{\mathcal{D}}_{k}$ which is before the thinning operation.  From the definition of KSD in \eqref{KSD_estimate}, we can write 
\begin{align}
        |\widetilde{\mathcal{D}}_k|^2\text{KSD}(\Lambda_{\widetilde{\mathcal{D}}_k})^2 &= \sum_{{h}_i\in \mathcal{D}_k}\sum_{h_j\in \mathcal{D}_k}  k_0({h}_i,{h}_j)
        \label{here_11}
        \\
        &=|{\mathcal{D}}_{k-1}|^2\text{KSD}(\Lambda_{\mathcal{D}_{k-1}})^2+\sum_{m=1}^{H/M}\bigg[k_0({h}_k^m,{h}_k^m)+  2\sum_{{h}_i\in {\mathcal{D}}_{k-1}}k_0({h}_i,{h}_k^m)\bigg]. 
        \label{here_12}
        \end{align}
        Next, for each $m$, since we select (SPMCMC based method in \cite{stein_point_Markov}) the sample ${h}_k^m$ from $\mathcal{Y}_m:=\{{h}_k^l\}_{l=1}^M$, and without loss of generality, we assume that $H/M$ is an integer. Now, from the SPMCMC based selection, we can write
        \begin{align}
            k_0({h}_k^m,{h}_k^m)+  2\sum_{{h}_i\in {\mathcal{D}}_{k-1}}k_0({h}_i,{h}_k^m) = & \inf_{{h}_k^m\in\mathcal{Y}_m} k_0({h}_k^m,{h}_k^m)+  2\sum_{{h}_i\in {\mathcal{D}}_{k-1}}k_0({h}_i,{h}_k^m)
            \nonumber
            \\
            \leq & S_k^2+2\inf_{{h}_k^m\in\mathcal{Y}_m} \sum_{{h}_i\in {\mathcal{D}}_{k-1}}k_0({h}_i,{h}_k^m). \label{inf_upper_bound}
        \end{align}
       The inequality in \eqref{inf_upper_bound} holds because we restrict our attention to  regions for which it holds that $k_0(\mathbf{x},\mathbf{x})\leq S_k^2$ for all $\mathbf{x}\in\mathcal{Y}_k^m$ for all $k$ and $m$. Utilizing the upper bound of \eqref{inf_upper_bound} into the right hand side of \eqref{here_12}, we get
        \begin{align}
        |\widetilde{\mathcal{D}}_k|^2\text{KSD}(\Lambda_{\widetilde{\mathcal{D}}_k})^2  &\leq|\mathcal{D}_{k-1}|^2\text{KSD}(\Lambda_{\mathcal{D}_{k-1}})^2+
        \frac{HS_k^2}{M}+2\sum_{m=1}^{H/M}\inf_{{h}_k^m\in\mathcal{Y}_m} \sum_{{h}_i\in {\mathcal{D}}_{k-1}}k_0({h}_i,{h}_k^m).
        \label{here_13}
\end{align}

Eqn \eqref{here_11} clearly differentiates our method from \cite{hawkins2022online} highlighting the novelty of our approach is deciphering the application of \text{KSD} to our model based RL problem. From the application of Theorem 5 \cite{hawkins2022online} for our formulation with $H$ new samples in the dictionary. 
\begin{align}
    2\inf_{{h}_k^m\in\mathcal{Y}_m} \sum_{{h}_i\in \widetilde{\mathcal{D}}_{k-1}}k_0({h}_i,{h}_k^m)\leq r_k\|f_k\|^2_{\mathcal{K}_0}+\frac{\text{KSD}(\Lambda_{\mathcal{D}_{k-1}})}{r_k}^2,
    \label{here_14}
\end{align}
for any arbitrary constant $r_k>0$. We use the upper bound in \eqref{here_14} to the right hand side of \eqref{here_13}, to obtain 
\begin{align}
         |\widetilde{\mathcal{D}}_k|^2\text{KSD}(\Lambda_{\widetilde{\mathcal{D}}_k})^2  \leq|\mathcal{D}_{k-1}|^2\left(1+\frac{H}{r_kM}\right) \text{KSD}(\Lambda_{\mathcal{D}_{k-1}})^2+\frac{HS_k^2+r_kH\|f_k\|_{\mathcal{K}_0}^2}{M}.
        \end{align}
        Next, we divide the both sides by $|\widetilde{\mathcal{D}}_k|^2=(|{\mathcal{D}_{k-1}}|+H/M)^2$ to obtain
        \begin{align}
        \text{KSD}(\Lambda_{\widetilde{\mathcal{D}}_k})^2 \leq \frac{|\mathcal{D}_{k-1}|^2}{\left(|\mathcal{D}_{k-1}|+H/M\right)^2}\left(1+\frac{H}{r_kM}\right)\text{KSD}(q_{\mathcal{D}_{k-1}})^2+\frac{H\left(S_k^2+r_k\|f_k\|_{\mathcal{K}_0}^2\right)}{M\left(|\mathcal{D}_{k-1}|+H/M\right)^2}\label{thinned}
\end{align}
Now, in this step we apply equation \eqref{eq: ksd goal} and replace $\Lambda_{\widetilde{\mathcal{D}}_k}$ with the thinned dictionary one $\Lambda_{{\mathcal{D}}_k}$ and rewriting equation \eqref{thinned} as 
\begin{align}
        \text{KSD}(\Lambda_{{\mathcal{D}}_k})^2 \leq \frac{|\mathcal{D}_{k-1}|^2}{\left(|\mathcal{D}_{k-1}|+H/M\right)^2}\left(1+\frac{H}{r_kM}\right)\text{KSD}(\Lambda_{{\mathcal{D}}_{k-1}})^2+\frac{H\left(S_k^2+r_k\|f_k\|_{\mathcal{K}_0}^2\right)}{M\left(|\mathcal{D}_{k-1}|+H/M\right)^2} + \epsilon_k.\label{recursion}
\end{align}
Note that we have established a recursive relationship for the KSD of the thinned distribution in \eqref{recursion}. This is quite interesting because it would pave the way to establish the regret result which is the eventual goal. After unrolling the recursion in \eqref{recursion}, we can write 
\begin{align}
 \text{KSD}(\Lambda_{{\mathcal{D}}_k})^2 \leq \sum_{i=1}^{k} \left( \frac{H\big(S_i^2+r_i\|f_i\|_{\mathcal{K}_0}^2\big)}{M\left(|\mathcal{D}_{i-1}|+H/M\right)^2}+ \epsilon_i \right)\left(\prod_{j=i}^{k-1} \frac{|\mathcal{D}_j|}{|\mathcal{D}_j|+H/M}\right)^2 \left( \prod_{j=i}^{k-1} \left(1+\frac{H}{r_{j+1}M}\right)\right).  \label{KSD_bound0}
\end{align}
Taking expectation on both sides of \eqref{KSD_bound0} and applying the log-sum exponential bound we get
\begin{align}
\label{eq: initial recursion with r 0}
\mathbb{E}\left[\text{KSD}(\Lambda_{{\mathcal{D}}_k})^2 \right] &
\leq \mathbb{E}\left[\exp\left(\frac{H}{M}\sum_{j=1}^k \frac{1}{r_j}\right)\sum_{i=1}^{k} \left( \frac{H(S_i^2+r_i\|f_i\|_{\mathcal{K}_0}^2)}{M\left(|\mathcal{D}_{i-1}|+H/M\right)^2}+ \epsilon_i \right)\left(\prod_{j=i}^{k-1} \frac{|\mathcal{D}_j|}{|\mathcal{D}_j|+H/M}\right)^2\right],
\end{align}
where the log-sum exponential bound is used as 
\begin{align}
    \label{eq: r product}
    \prod_{j=i}^{k-1}\left(1+\frac{H}{r_{j+1}M}\right)\leq \exp\left(\frac{H}{M}\sum_{j=1}^n \frac{1}{r_j}\right).
\end{align}
Next, we consider the inequality in \eqref{eq: initial recursion with r 0}, ignoring the exponential term, we further decompose the remaining summation term into two parts, respectively, as:
\begin{align}
\mathbb{E}&\left[\sum_{i=1}^{k} \left( \frac{H(S_i^2+r_i\|f_i\|_{\mathcal{K}_0}^2)}{M\left(|\mathcal{D}_{i-1}|+H/M\right)^2}+ \epsilon_i \right)\left(\prod_{j=i}^{k-1} \frac{|\mathcal{D}_j|}{|\mathcal{D}_j|+H/M}\right)^2\right]
\label{eq: T1 T2 T3 introduction}
\\
&=\underbrace{\mathbb{E}\left[\sum_{i=1}^{k} \left( \frac{H(S_i^2+r_i\|f_i\|_{\mathcal{K}_0}^2)}{M\left(|\mathcal{D}_{i-1}|+H/M\right)^2}\right)\left(\prod_{j=i}^{k-1} \frac{|\mathcal{D}_j|}{|\widetilde{D}_j|+H/M}\right)^2\right]}_{T_1}
+
\underbrace{\mathbb{E}\left[\sum_{i=1}^{k} \epsilon_i\left(\prod_{j=i}^{k-1} \frac{|\mathcal{D}_j|}{|\mathcal{D}_j|+H/M}\right)^2\right]}_{T_2}\nonumber,
\end{align}

\begin{figure}[t]
     \centering
     \begin{subfigure}[b]{0.3\textwidth}
         \centering
\includegraphics[width=\textwidth]{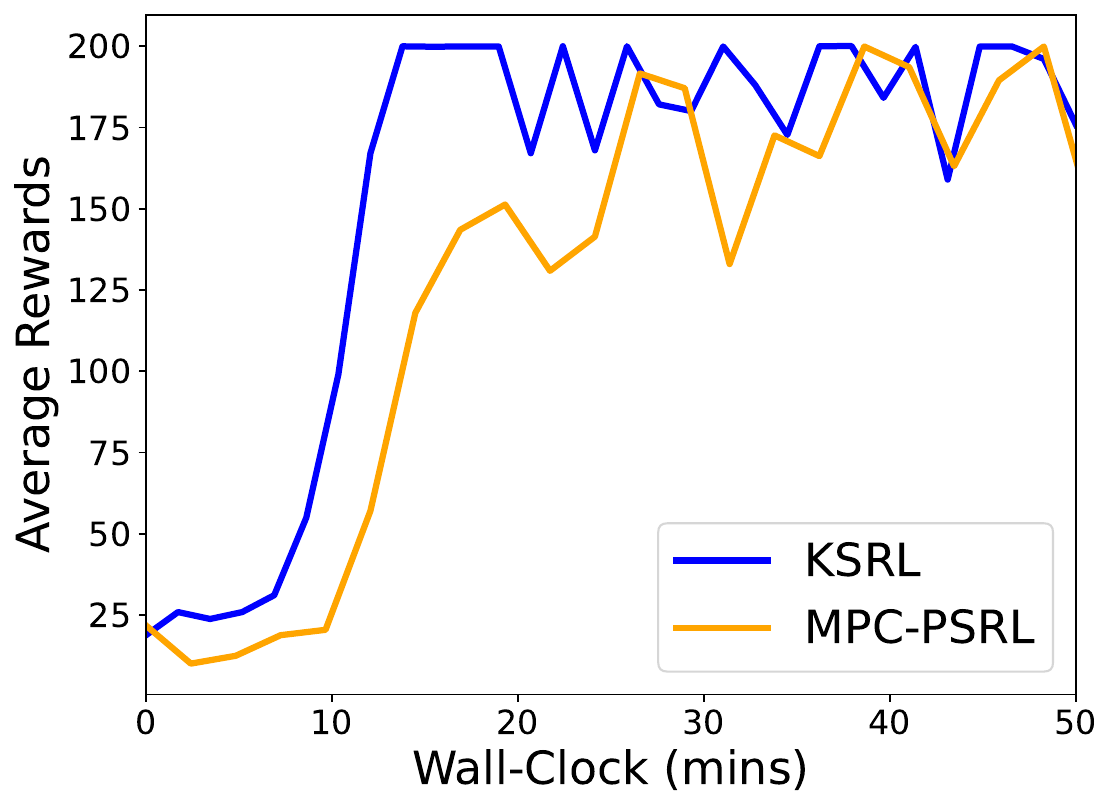}
         \caption{}
         \label{fig:wall clock1}
     \end{subfigure}
          \hfill
     \begin{subfigure}[b]{0.3\textwidth}
         \centering
\includegraphics[width=\textwidth]{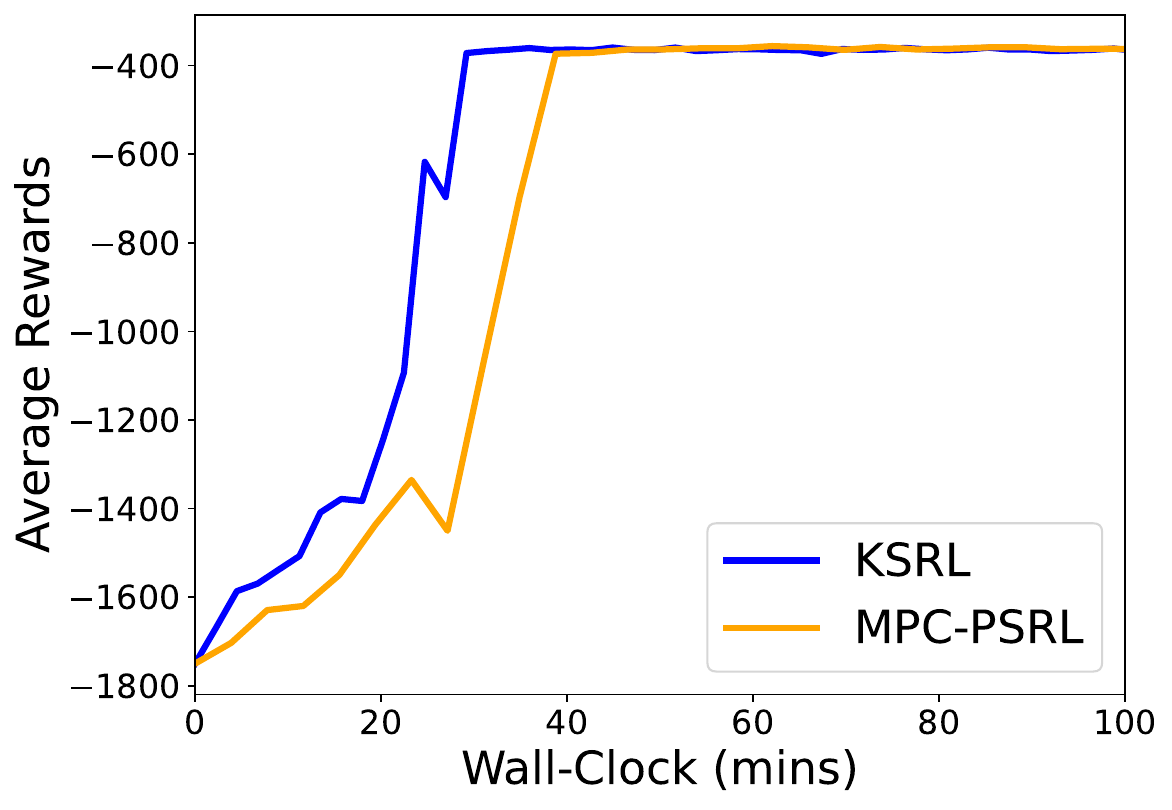}
         \caption{}
         \label{fig:wall clock2}
     \end{subfigure}
     \hfill
          \begin{subfigure}[b]{0.3\textwidth}
         \centering
\includegraphics[width=\textwidth]{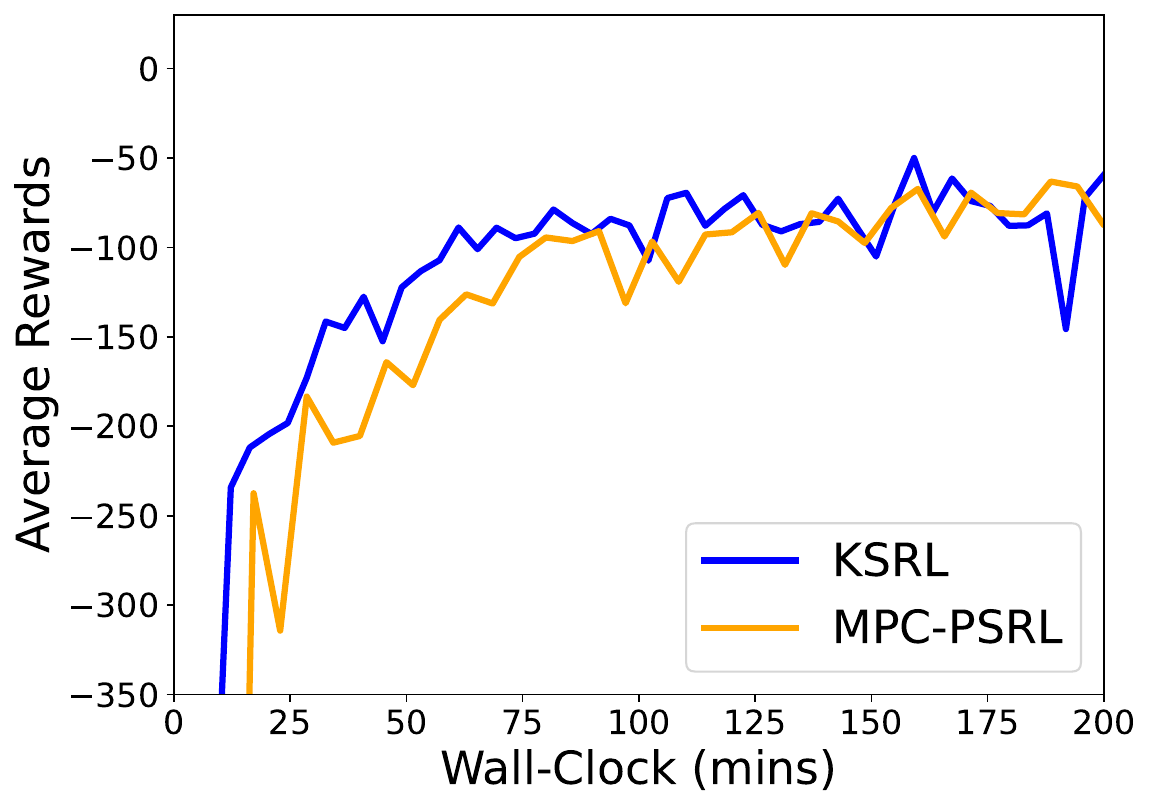}
         \caption{}
         \label{fig:wall clock3}
     \end{subfigure}\\
          \hfill
     \begin{subfigure}[b]{0.3\textwidth}
         \centering
\includegraphics[width=\textwidth]{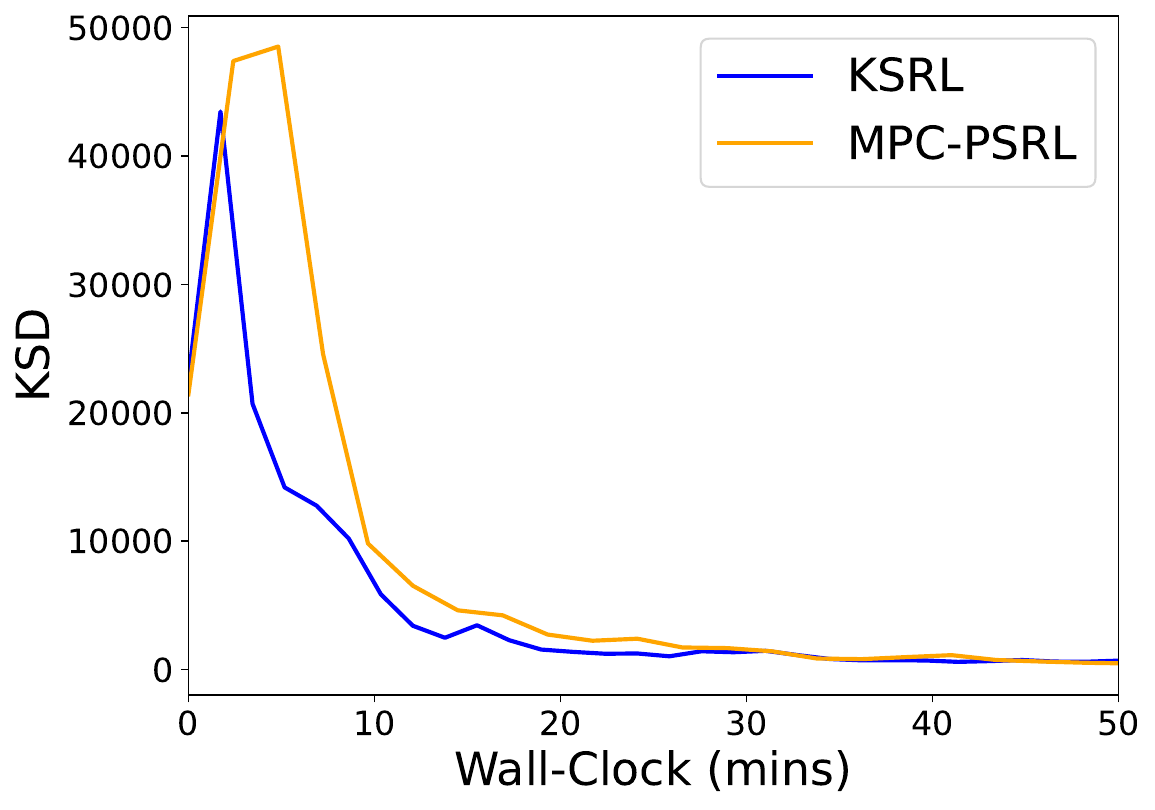}
         \caption{}
         \label{fig:wall clock4}
     \end{subfigure}
          \hfill
          \begin{subfigure}[b]{0.3\textwidth}
         \centering
\includegraphics[width=\textwidth]{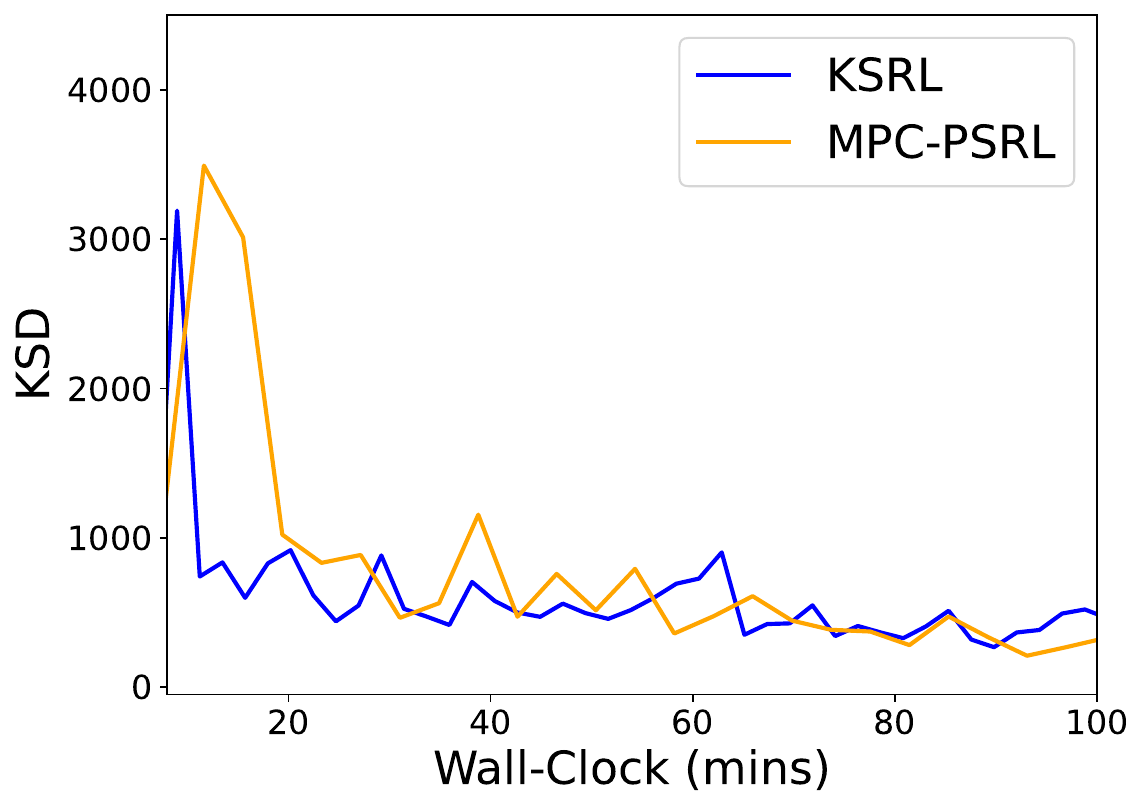}
         \caption{}
         \label{fig:wall clock5}
     \end{subfigure}
          \hfill
     \begin{subfigure}[b]{0.3\textwidth}
         \centering
\includegraphics[width=\textwidth]{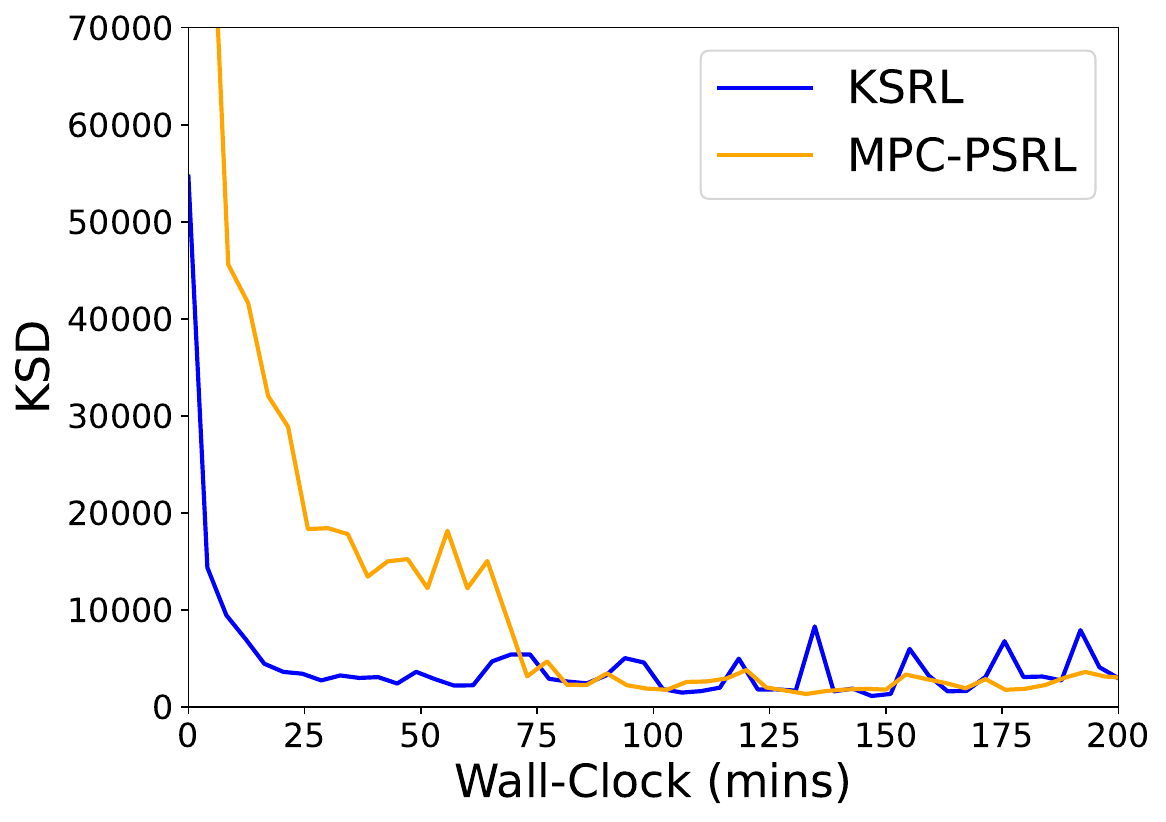}
         \caption{}
         \label{fig:wall clock6}
     \end{subfigure}
        \caption{Performance against wall clock time and KSD Convergence results: \textbf{(a)-(c)} shows the average reward return against wall clock time (in minutes) for modified Cartpole, Pendulum, and Reacher (mean across 5 runs). \textbf{(d)-(f)} provides the evidence of KSD convergence which shows we are learning the target posterior effectively without any bias (even we are compressing the dictionary). From \textbf{(a)-(c)} it is evident that \algo (blue) is able to achieve the similar performance even earlier than the existing dense counterparts with no thinning. Wall-clock time measured in minutes for runs in CPU.}
        \label{fig:three graphs2}
\end{figure}

where $T_1$ corresponds to the sampling error and $T_2$ corresponds to the error due to the proposed thinning scheme. The term $T_1$ represents the bias incurred at each step of the un-thinned point selection scheme. The term $T_2$ is the bias incurred by the thinning operation carried out at each step. However, for our case in the model-based reinforcement learning setting, the bias term $T_1$ will be different in our case as the samples are generated in a sequential manner in each episode from a Transition kernel with a decaying budget. Let develop upper bounds on both $T_1$ and $T_2$ as follows. 

\textbf{Bound on $T_1$:} Let consider the first term on the right hand side of \eqref{eq: T1 T2 T3 introduction} as follow
\begin{align}
     T_1 =&\mathbb{E}\left[\sum_{i=1}^{k} \left( \frac{H(S_i^2+r_i\|f_i\|_{\mathcal{K}_0}^2)}{M\left(|\mathcal{D}_{i-1}|+H/M\right)^2}\right)\left(\prod_{j=i}^{k-1} \frac{|\mathcal{D}_j|}{|\mathcal{D}_j|+H/M}\right)^2\right] 
     \nonumber
     \\
     =&\mathbb{E}\left[\sum_{i=1}^{k} \left( \frac{H(S_i^2+r_i\|f_i\|_{\mathcal{K}_0}^2)}{M\left(|\mathcal{D}_{k-1}|+H/M\right)^2}\right)\left(\prod_{j=i}^{k-1} \frac{|\mathcal{D}_j|}{|\mathcal{D}_{j-1}|+H/M}\right)^2\right] \label{here1111},
\end{align}
where the equality in \eqref{here1111} holds by rearranging the denominators in the multiplication, and pulling $\left(|\mathcal{D}_{k-1}|+H/M\right)^2$ inside the first term. Next, from the fact that ${|\mathcal{D}_j|}\leq {|\mathcal{D}_{j-1}|+H/M}$ which implies that the product will be less that $1$, we can upper bound the right hand side of \eqref{here1111} as follows
\begin{align}\label{eq: bound computation0}
    T_1 &\leq\mathbb{E}\left[\sum_{i=1}^{k} \left( \frac{H(S_i^2+r_i\|f_i\|_{\mathcal{K}_0}^2)}{M\left(|\mathcal{D}_{k-1}|+H/M\right)^2}\right)\right]\\
    & =\sum_{i=1}^{k} \left( \frac{H(S_i^2+r_i\mathbb{E}\left[\|f_i\|_{\mathcal{K}_0}^2\right])}{M\left(|\mathcal{D}_{k-1}|+H/M\right)^2}\right),\label{here_2222}
\end{align}
where we took expectation inside the summation and apply it to the random variable in the numerator. From the model order growth condition, we note that $|\mathcal{D}_{k-1}|+H/M \geq |{\mathcal{D}_k}| \geq f(k)$, which implies that $1/\left({|{\mathcal{D}_{k-1}}|+H/M}\right)^2\leq 1/f(k)^2$, which we utilize in the right hand side of  \eqref{here_2222} to write
\begin{align}\label{eq: bound computation}
    T_1 &\leq\mathbb{E}\left[\sum_{i=1}^{k} \left( \frac{H(S_i^2+r_i\|f_i\|_{\mathcal{K}_0}^2)}{M\left(|\mathcal{D}_{k-1}|+H/M\right)^2}\right)\right]\\
    & =\sum_{i=1}^{k} \left( \frac{H(S_i^2+r_i\mathbb{E}\left[\|f_i\|_{\mathcal{K}_0}^2\right])}{M f(k)^2}\right)\label{here_22222}
\end{align}

Following the similar logic mentioned in  \cite[Appendix A, Eqn. (17)]{stein_point_Markov}, we can upper bound $\mathbb{E}\left[\|f_i\|_{\mathcal{K}_0}^2\right]$ as
\begin{align}
    \mathbb{E}\left[\|f_i\|_{\mathcal{K}_0}^2\right] \leq \frac{4b}{\gamma} \exp\left(-\frac{\gamma}{2}S_i^2\right) + \frac{4}{H/M}S_i^2, \label{intermediate}
\end{align}
which is based on the assumption that $\mathbb{E}_{\mathbf{y}\sim P}\left[\exp(\gamma \kappa_{0}(\mathbf{y},\mathbf{y}))\right]=b<\infty$. This implies that
\begin{align}
    T_1 & \leq \frac{1}{ f(k)^2}\sum_{i=1}^{k} \left( {(H/M+4r_i)S_i^2+\frac{4br_i H}{\gamma M} \exp\left(-\frac{\gamma}{2}S_i^2\right))}\right)\label{here_2222222},
\end{align}
which we obtain by applying the upper bound in \eqref{intermediate} into \eqref{here_22222}.
After simplification, we can write
\begin{align}
    T_1 & \leq \mathcal{O} \left(\frac{k\log(k)}{f(k)^2}\right)\label{here_2222222222},
\end{align}
which provide the bound on $T_1$. Next, we derive the bound on $T_2$ as follows. 
\paragraph{Bound on $T_2 :$} Let us consider the term $T_2$ from \eqref{eq: T1 T2 T3 introduction} as follows
\begin{align}
    T_2=\mathbb{E}\left[\sum_{i=1}^{k} \epsilon_i\left(\prod_{j=i}^{k-1} \frac{|\mathcal{D}_j|}{|\mathcal{D}_j|+H/M}\right)^2\right].\label{T_1_bound}
\end{align}
This term is extra in the analysis and appears due to the introduction of compression budget $\epsilon_k$ into the algorithm. We need control the growth of this term, and by properly designing $\epsilon_k$, we need to make sure $T_2$ goes to zero at least as fast at $T_1$ to obtain a sublinear regret analysis. We start by observing that $|\mathcal{D}_j|\leq (H/M)j$ which holds trivially and hence implies
\begin{align}\label{dictionary_inequality}
    \frac{|\mathcal{D}_j|}{|\mathcal{D}_{j+1}|}\leq \frac{j}{j+1}. 
\end{align}
From the algorithm construction, we know that $|\mathcal{D}_{j+1}|\leq |\mathcal{D}_j|+H/M$, applying this to \eqref{dictionary_inequality}, we obtain
\begin{align}\label{dictionary_inequality2}
    \frac{|\mathcal{D}_j|}{|\mathcal{D}_j|+H/M}\leq \frac{j}{j+1}. 
\end{align}
Utilize the upper bound in \eqref{dictionary_inequality2} to the right hand side of \eqref{T_1_bound}, to write
\begin{align}
    T_2\leq \sum_{i=1}^{k} \epsilon_i\left(\prod_{j=i}^{k-1} \frac{j}{j+1}\right)^2=\sum_{i=1}^{k} \epsilon_i\frac{i^2}{k^2} =\frac{1}{k^2}\sum_{i=1}^{k} \epsilon_i{i^2} .\label{T_1_bound_extra}
\end{align}
The above bound implies that we should choose the compression budget $\epsilon_i$ such that $T_2$ goes to zero at least as fast at $T_1$ 

\begin{align}
     \frac{1}{k^2}\sum_{i=1}^{k} \epsilon_i{i^2} \leq \frac{k\log(k)}{f(k)^2}\\
     \sum_{i=1}^{k} \epsilon_i{i^2} \leq \frac{k^3 \log(k)}{f(k)^2}
     .\label{T_1_bound_2}
\end{align}

To satisfy the above condition, we choose $\epsilon_{i}=\frac{\log(i)}{f(i)^2}$, and obtain
\begin{align}
  \sum_{i=1}^{k} \epsilon_i{i^2}= \sum_{i=1}^{k} \frac{i^2\log(i)}{f(i)^2}
  \nonumber
  \\
  \leq \sum_{i=1}^{k} \frac{k^2\log(k)}{f(k)^2}\nonumber
  \\
  \leq \frac{k^3\log(k)}{f(k)^2},\label{T_1_bound_4}
\end{align}
which satisfy the upper bound in \eqref{T_1_bound_2}, which shows that $\epsilon_{i}=\frac{\log(i)}{f(i)^2}$ is a valid choice. Hence, $T_2\leq  \frac{k\log(k)}{f(k)^2}$. 
\begin{figure}
     \centering
     \begin{subfigure}[b]{0.3\textwidth}
         \centering
\includegraphics[width=\textwidth]{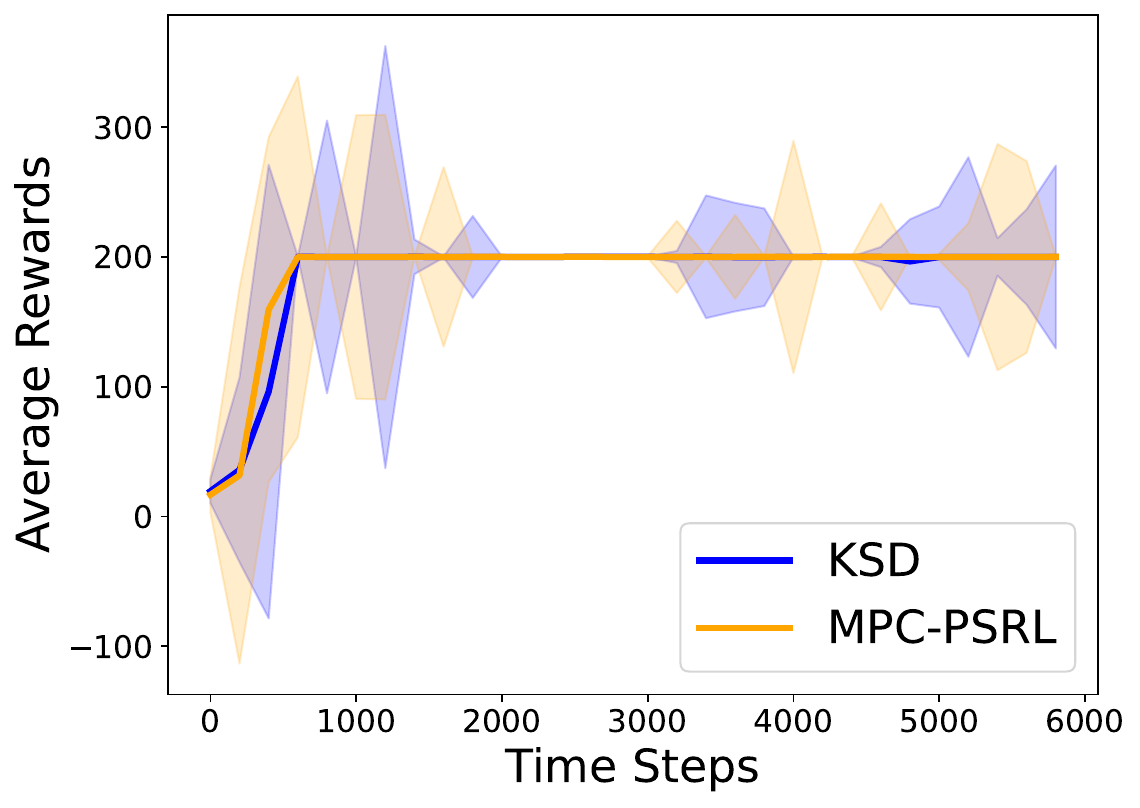}
         \caption{.}
         \label{fig:with rewards1}
     \end{subfigure}
          \hfill
     \begin{subfigure}[b]{0.3\textwidth}
         \centering
\includegraphics[width=\textwidth]{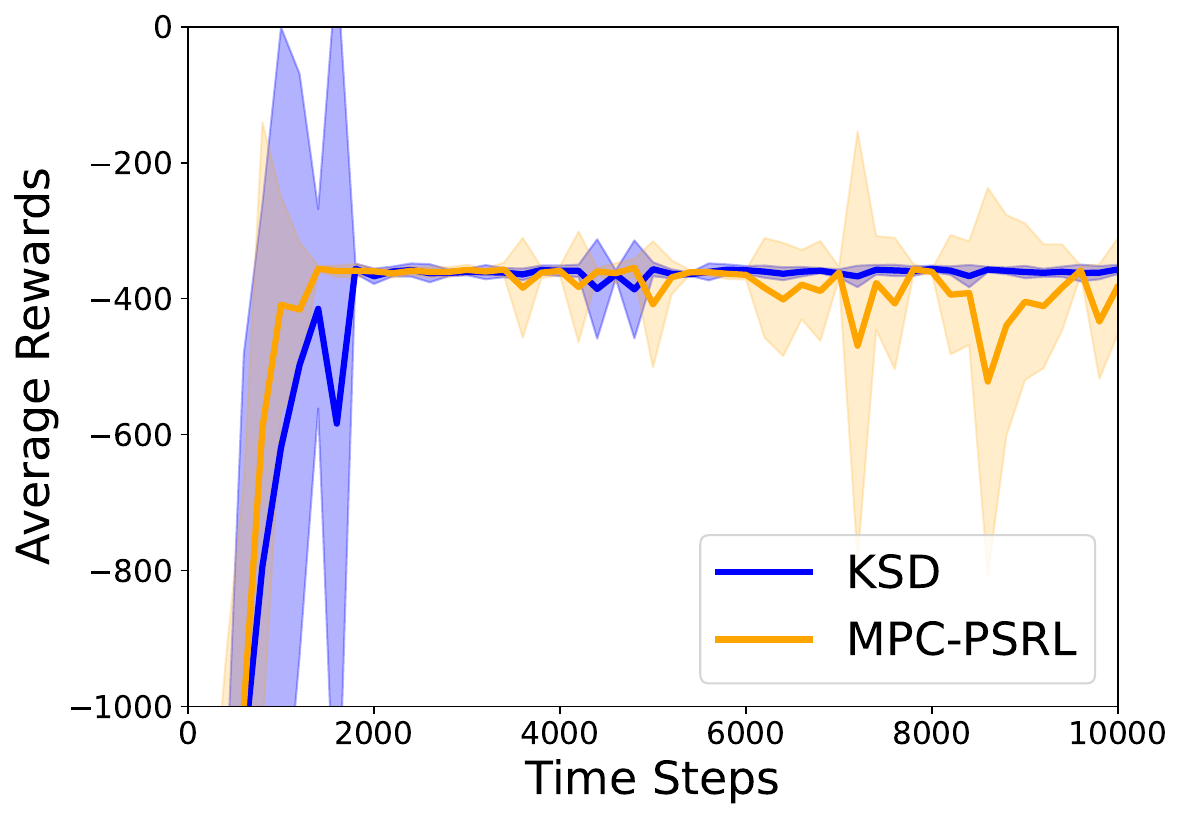}
         \caption{}
         \label{fig:with rewards2}
     \end{subfigure}
     \hfill
          \begin{subfigure}[b]{0.3\textwidth}
         \centering
\includegraphics[width=\textwidth]{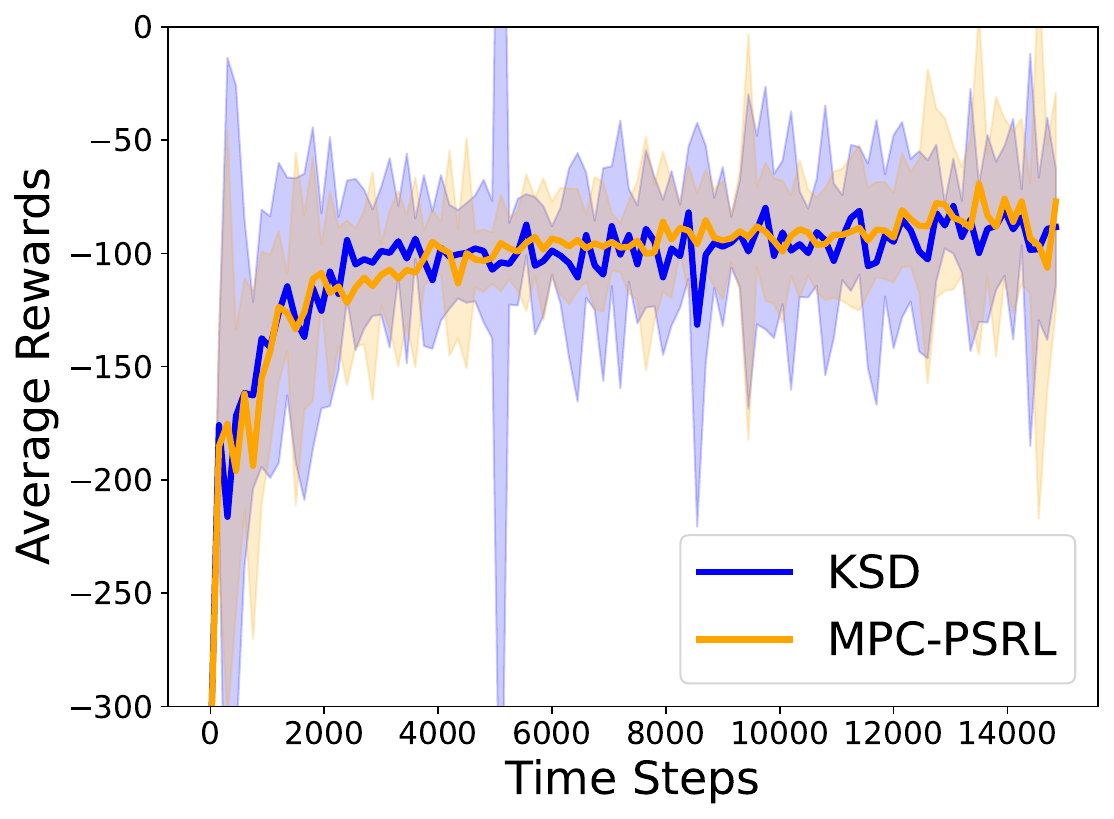}
         \caption{}
         \label{fig:with rewards3}
     \end{subfigure}\\

          \hfill
     \begin{subfigure}[b]{0.3\textwidth}
         \centering
\includegraphics[width=\textwidth]{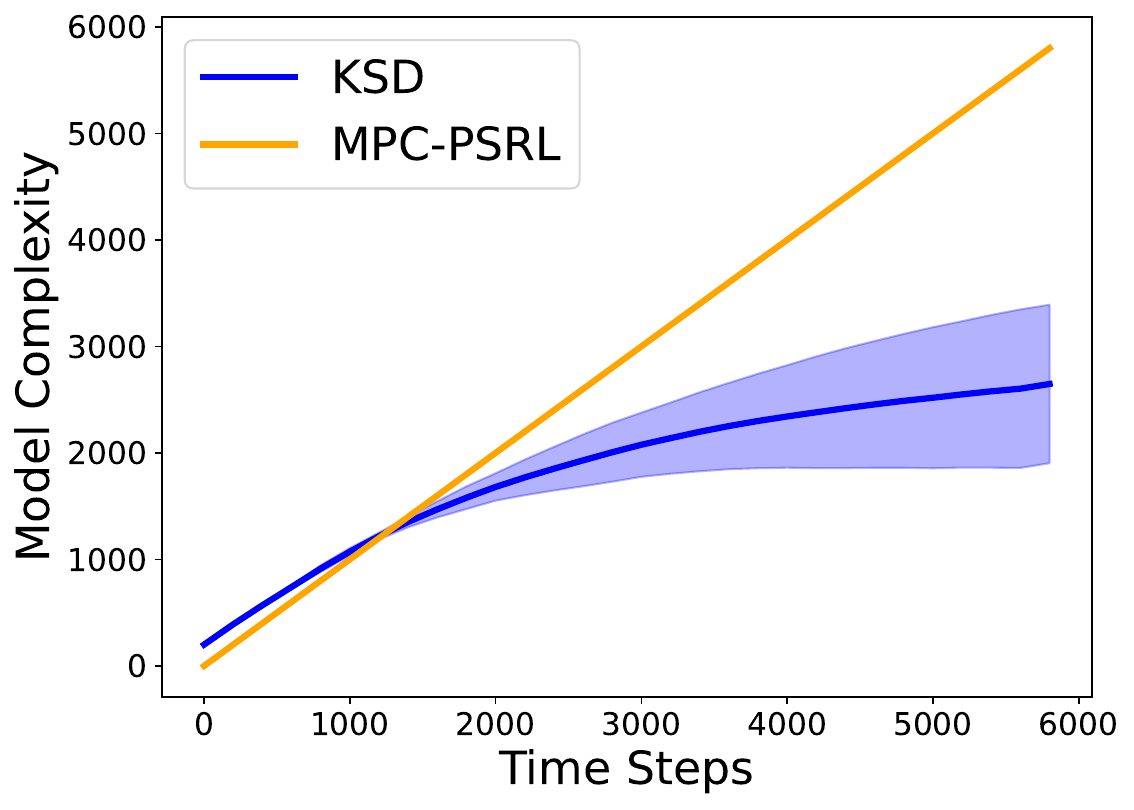}
         \caption{}
         \label{fig:with rewards4}
     \end{subfigure}
          \hfill
          \begin{subfigure}[b]{0.3\textwidth}
         \centering
\includegraphics[width=\textwidth]{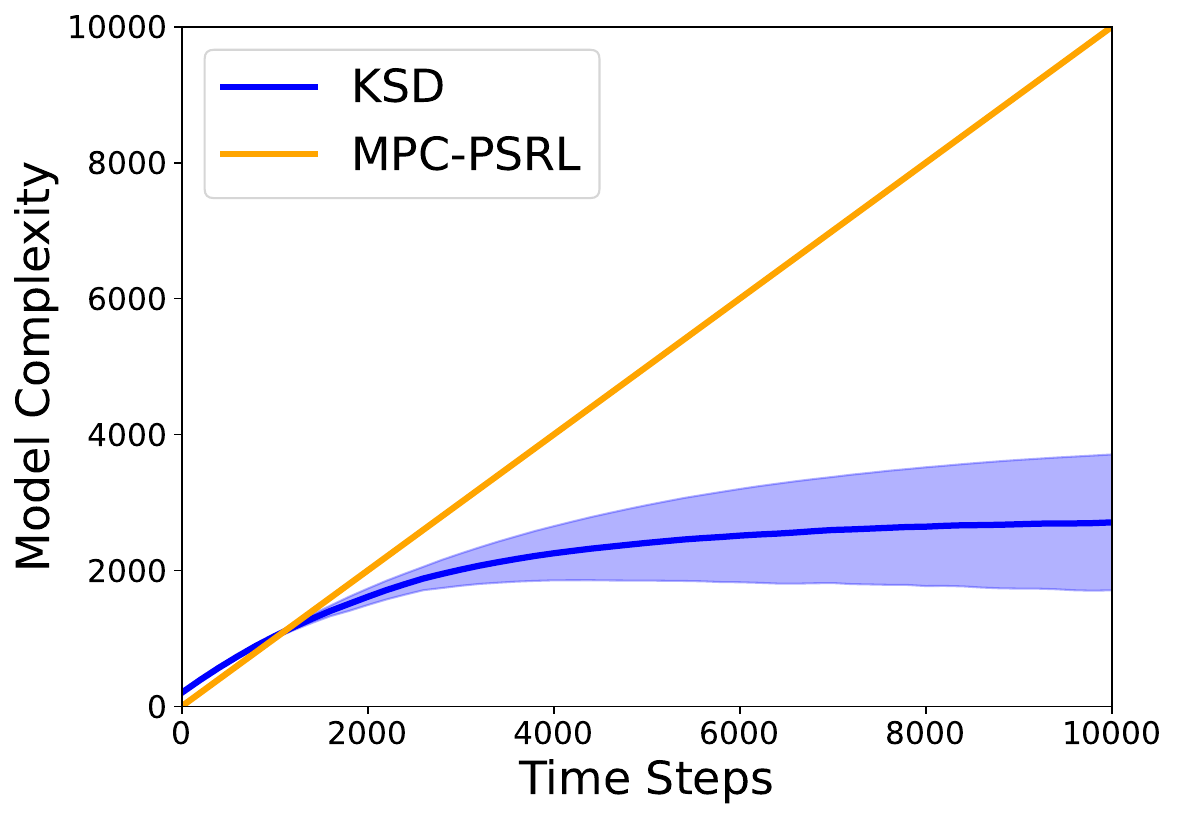}
         \caption{}
         \label{fig:with rewards5}
     \end{subfigure}
          \hfill
     \begin{subfigure}[b]{0.3\textwidth}
         \centering
\includegraphics[width=\textwidth]{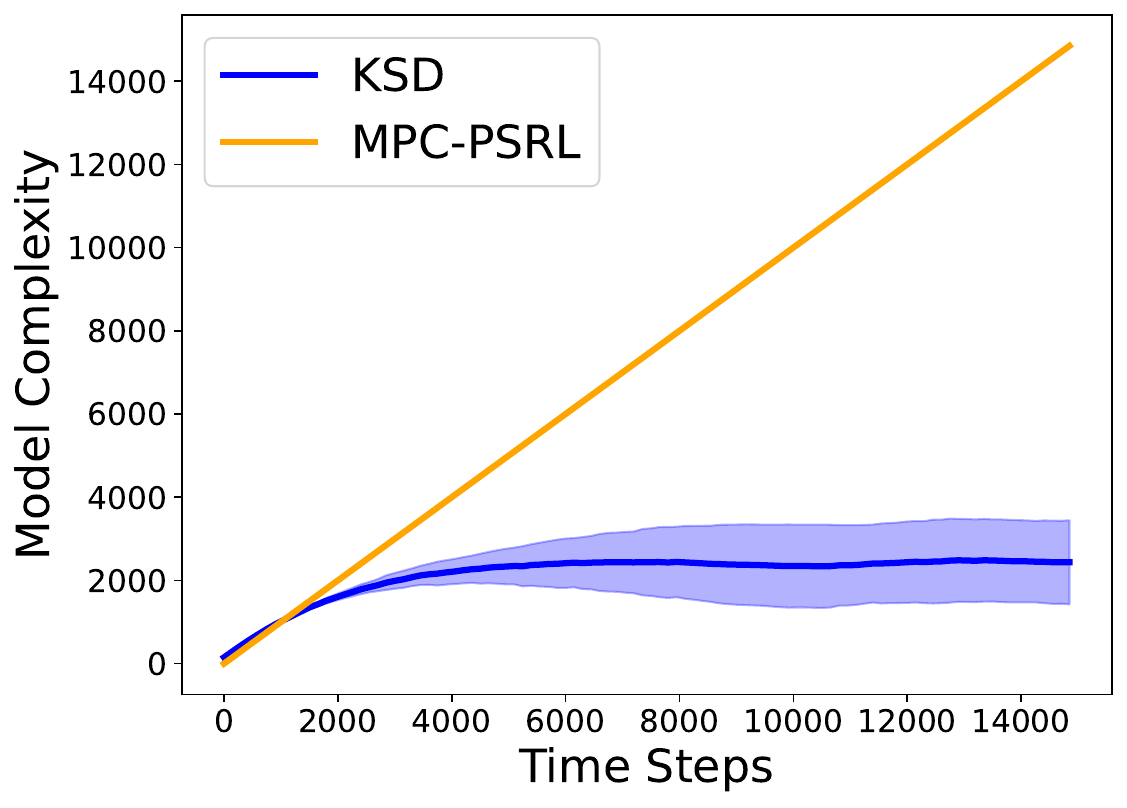}
         \caption{}
         \label{fig:with rewards6}
     \end{subfigure}
        \caption{\textbf{(a)-(c)} compares the average cumulative reward return achieved by the proposed {\algo} (shown in blue) algorithm with MPC-PSRL \cite{pmlr-v139-fan21b}, SAC \cite{haarnoja2018soft}, and DDPG \cite{barth2018distributed} for \textbf{Cartpole}, \textbf{Pendulum}, and \textbf{Pusher} with oracle rewards.  \textbf{(d)-(f)} compares the model-complexity.  We note that {\algo} is able to achieve the maximum average reward at-par with the current state of the art MPC-PSRL with drastically reduced model complexity. Solid curves represent the average across five trials (seeds), shaded areas correspond to the standard deviation amongst the trials}.
        \label{fig:with_rewards}
\end{figure}

Finally, after substituting the upper bounds for $T_1$ and $T_2$ into \eqref{eq: T1 T2 T3 introduction}, we obtain
\begin{align}
\mathbb{E}&\left[\sum_{i=1}^{k} \left( \frac{H(S_i^2+r_i\|f_i\|_{\mathcal{K}_0}^2)}{M\left(|\mathcal{D}_{i-1}|+H/M\right)^2}+ \epsilon_i \right)\left(\prod_{j=i}^{k-1} \frac{|\mathcal{D}_j|}{|\mathcal{D}_j|+H/M}\right)^2\right]
\label{eq: T1 T2 T3 introduction2}\leq \frac{2k\log(k)}{f(k)^2}.
\end{align}
Now revisiting the inequality in \eqref{KSD_bound0}, we write
\begin{align}
\mathbb{E}\left[\text{KSD}(\Lambda_{{\mathcal{D}}_k})^2 \right] &
\leq \mathbb{E}\left[\exp\left(\frac{H}{M}\sum_{j=1}^k \frac{1}{r_j}\right)\sum_{i=1}^{k} \left( \frac{H(S_i^2+r_i\|f_i\|_{\mathcal{K}_0}^2)}{M\left(|\mathcal{D}_{i-1}|+H/M\right)^2}+ \epsilon_i \right)\left(\prod_{j=i}^{k-1} \frac{|\mathcal{D}_j|}{|\mathcal{D}_j|+H/M}\right)^2\right]
\nonumber
\\
\leq & \exp\left(\frac{H}{M}\sum_{j=1}^k \frac{1}{r_j}\right) \left(\frac{2k\log(k)}{f(k)^2}\right)\nonumber
\\
\leq & e \left(\frac{2k\log(k)}{f(k)^2}\right).
\label{eq: initial recursion with r_2}
\end{align}
where the last equality holds by the selection $r_j=\frac{Hk}{M}$ for all $j$. Finally, we collect all the constants in $C$ to write
\begin{align}
\mathbb{E}\left[\text{KSD}(\Lambda_{{\mathcal{D}}_k})^2 \right] 
\leq & \mathcal{O}\left(\frac{k\log(k)}{f(k)^2}\right).
\label{eq: initial recursion with r_222}
\end{align}
Next, from applying Jensen's inequality on the left hand side in \eqref{eq: initial recursion with r_222}, and we can write
\begin{align}
\mathbb{E}\left[\text{KSD}(\Lambda_{\widetilde{D}_k})\right] 
\leq & \mathcal{O}\left(\frac{\sqrt{k\log(k)}}{f(k)}\right).
\label{eq: initial recursion with r_2222}
\end{align}

Hence proved.
\end{proof}
\section{Proof of Theorem \ref{thm: decaying budget}}\label{proof_theorem_1}
\begin{proof}
Based on the equality in \eqref{intermediate}, the first part of our regret in \eqref{regret} becomes zero and we only need to derive  and upper bound for $\Delta^{II}_k$. Recall that, we have
\begin{align}
  \Delta^{II}_k=  \int \rho(s_1)(  V_{\tilde{\mu}^k}^{\widetilde{{M}}^k}(s_1)  -  V_{\tilde{\mu}^k}^{{M^*}}(s_1)   ) ds_1.
\end{align}

Following the Bellman equations representation of value functions, We can factorize $\Delta^{II}_k$ in-terms of the immediate rewards $r$ and the future value function $U$ (cf. \cite{osband2014model}) as follows
\begin{align}\label{refret_definition_theorem}
 \mathbb{E}[\Delta^{II}_k|\mathcal{D}_{k}]=&\mathbb{E}[{\Delta}_k(r)+{\Delta}_k(f)|\mathcal{D}_{k}],
 \end{align}
 where we define
 \begin{align}
 {\Delta}_k(r) = &\sum_{i=1}^H ({r}^k(\hat{h}_i)- {r}^*(\hat{h}_i)),\\
 {\Delta}_k(P) = &\sum_{i=1}^H (U_{i}^k({P}^k(\hat{h}_i))- U_{i}^k({P}^*(\hat{h}_i))).
\end{align}
%
% The primary objective is thus to generate an efficient representation $\widetilde{D}^{k}$ from the unthinned dictionary after the $k^{th}$ episode $D_k$ such that the KSD squared distance between the compressed and uncompressed posterior distribution is $\epsilon$ bounded.

% \begin{equation}
%     \label{eq: ksd goal}
%     KSD(\widetilde{P}^k)^2 <KSD(P^k)^2+\epsilon.
% \end{equation}

% \begin{lemma}[Upper Bound on Kernel Stein Discrepancy for PSRL]\label{thm: decaying budget}
% At the $k^{th}$ episode, the thinned transition model is $\widetilde{P}^k$, unthinned history/dictionary $D_k$ then the KSD at the $k^{th}$ episode can be upper-bounded as 

% \begin{equation}
% \label{eq: decaying budget conclusion}
% \mathbb{E}[KSD(\widetilde{P}^k)^2] \leq C\frac{k\log(k)}{f(k)^2}.
% \end{equation}
% \end{lemma}

% under the assumptions that the kernel $k_0$ satisfies $\mathbb{E}_{{\gamma}\sim P}\left[e^{\beta k_0({y},{y})}\right]=b < \infty$, dictionary sizes are lower bounded as $|\mathcal{D}_k| \geq Cf(k)$ where $f(k)= o\left(\sqrt{k\log(k)}\right)$ and specify the compression budget $\epsilon_k = \log(k)/f(k)^2$. 
% \gamma_k

% Next we try to prove the convergence with regards to KSD metric in the subsequent steps.
% Before starting our convergence analysis, we introduce 2 standard requirements on the transition model $\widetilde{P}$ given in sections 5.1.1 and 5.1.2. We assume that the transition model $\widetilde{P}$ doesn't possess a heavy tail and the accepted proposals have low norm.

Now, for our Stein based thinning algorithm, we thin the updated dictionary $\widetilde{\mathcal{D}}_{k}$ after every episode ($H$ steps) and obtain a compressed and efficient representation of dictionary given by ${\mathcal{D}}_{k}$. Let us derive the upper bound on $\mathbb{E}[{\Delta}_k(P)]$as follows. 
The Bayes regret at the $k^{th}$ episode can be written as 
\begin{align}\label{lipschitz_upper}
    \mathbb{E}[{\Delta}_k(P)] &= \mathbb{E} \left[\sum_{i=1}^H (U_{i}^k({P}^k(\hat{h}_i))- U_{i}^k({P}^*(\hat{h}_i)))\right].
    \end{align}
Utilize the upper bound from the statement of Lemma \ref{lemma_1} to write
\begin{align}
    \mathbb{E}[{\Delta}_k(P)] &\leq \mathbb{E} \left[\sum_{i=1}^H d HR_{\text{max}} \text{KSD}({P}^k(\cdot|\hat{h}_i))\right]\nonumber
\\
&= d HR_{\text{max}}\sum_{i=1}^H \mathbb{E}\left[\text{KSD}(\tilde{P}^k(h_i))\right]. \label{KSD_bound}
    \end{align}
 This is a an important step and point of departure from the existing state of the art regret analysis for model based RL methods \cite{pmlr-v139-fan21b,osband2014model}. Instead of utilizing the naive upper bound of Total Variation distance in the right hand side of \eqref{lipschitz_upper}, we follow a different approach and bound it via the Kernel Stein Discrepancy which is a novel connection explored for the first time in this work. {We remark that the KSD upper bound in Lemma \ref{lemma_2} is for the joint posterior where samples are $h_i=(s,a,s')$. We note here that since the score function if independent of the normalizing constant, we can write $\nabla \log \tilde P^k(\cdot|h_i')=\nabla \log \tilde P^k(h_i)$, therefore we utilize the KSD upper bound of joint posterior in Lemma \ref{lemma_2} on the KSD term per episode in the right hand side of \eqref{KSD_bound} to obtain}
\begin{align}
    \mathbb{E}[{\Delta}_k(P)] 
    & \leq d HR_{\text{max}} \sum_{i=1}^H \frac{\sqrt{k \log(k)}}{f(k)} = d H^2 R_{\text{max}} \frac{\sqrt{k \log(k)}}{f(k)}.
\end{align}
Next, we take summation over the number of episodes which are given by $[\frac{T}{H}]$ where $T$ is the total number of time steps in the environments, and $H$ is the episode length. So, after summing over $k=1$ to $[\frac{T}{H}]$, we obtain
\begin{align}\label{here_1}
    \sum_{k=1}^{[\frac{T}{H}]}\mathbb{E}[{\Delta}_k(P)]  \leq d H^2 R_{\text{max}} \sum_{k=1}^{[\frac{T}{H}]} \frac{\sqrt{k \log(k)}}{f(k)}.
\end{align}
To derive the explicit regret rates, we assume our dictionary growth function $f(k) = \sqrt{k^{\alpha+1} \log(k)}$ with range of $\alpha \in [0,1]$. Substituting this into \eqref{here_1}, we can write 
\begin{align}
  \sum_{k= 1}^{\frac{T}{H}} \frac{\sqrt{k \log(k)}}{f(k)}= \sum_{k= 1}^{\frac{T}{H}} \frac{\sqrt{k \log(k)}}{\sqrt{k^{\alpha+1} \log(k)}}= \sum_{k= 1}^{\frac{T}{H}} k^{-\frac{\alpha}{2}} \leq \int_{0}^{\frac{T}{H}} x^{-\frac{\alpha}{2}} dx = \frac{2}{1-\alpha/2} T^{1-\frac{\alpha}{2}} H^{1+\frac{\alpha}{2}}. \label{integral_bound} %
\end{align}

\begin{figure}[t]
     \centering
     \begin{subfigure}[b]{0.4\textwidth}
         \centering
\includegraphics[width=\textwidth]{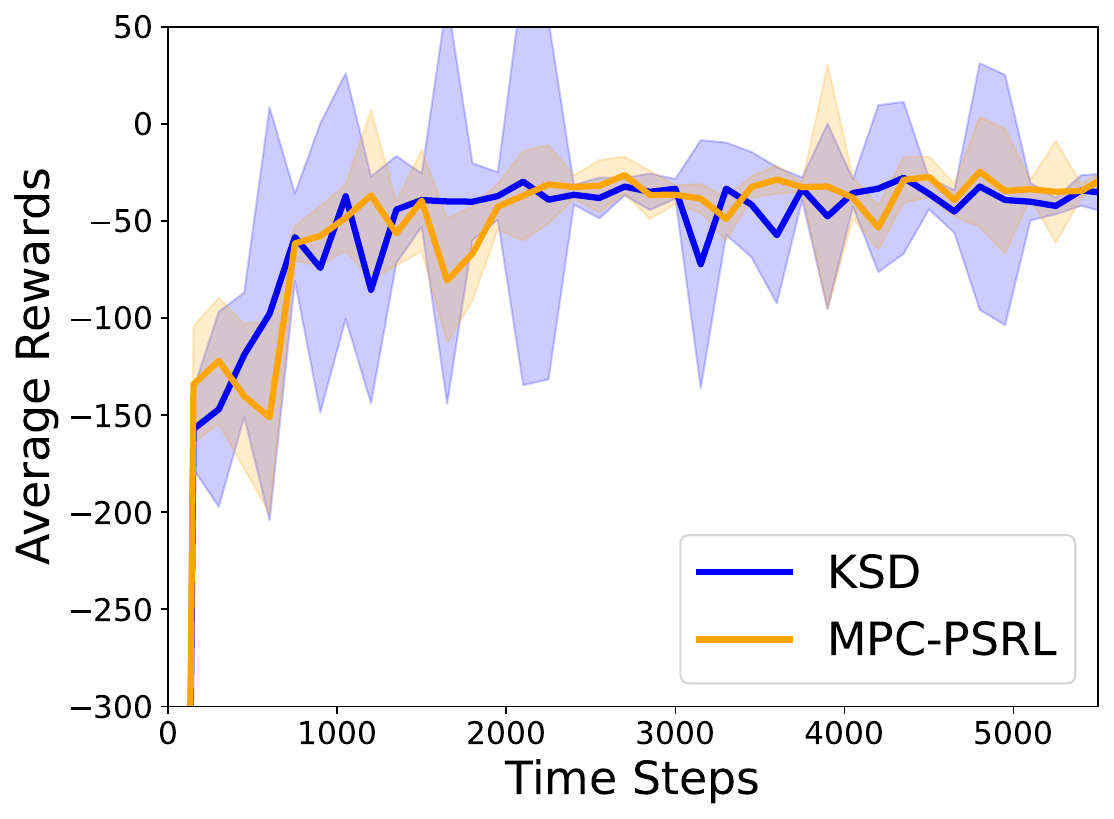}
         \caption{.}
         \label{fig:reacher1}
     \end{subfigure}
    %%      \hfill
     \begin{subfigure}[b]{0.4\textwidth}
         \centering
\includegraphics[width=\textwidth]{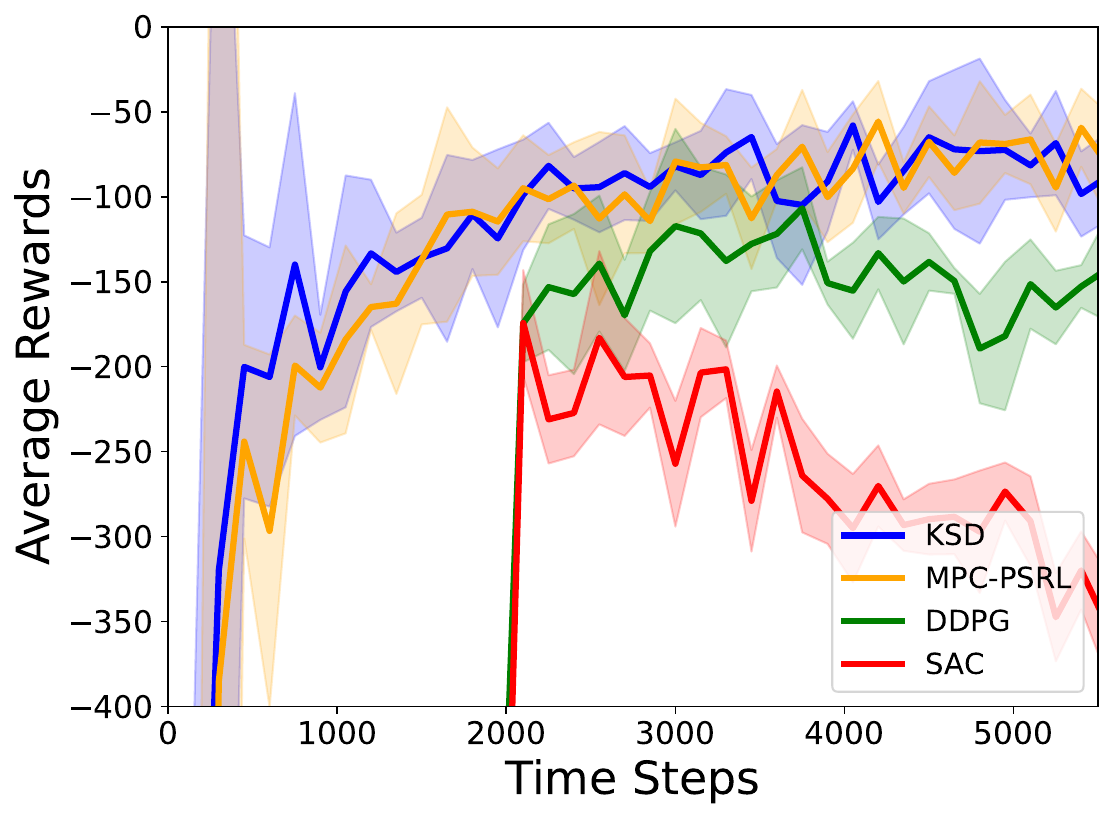}
         \caption{}
         \label{fig:reacher2}
     \end{subfigure}
  %   \hfill
          \begin{subfigure}[b]{0.4\textwidth}
         \centering
\includegraphics[width=\textwidth]{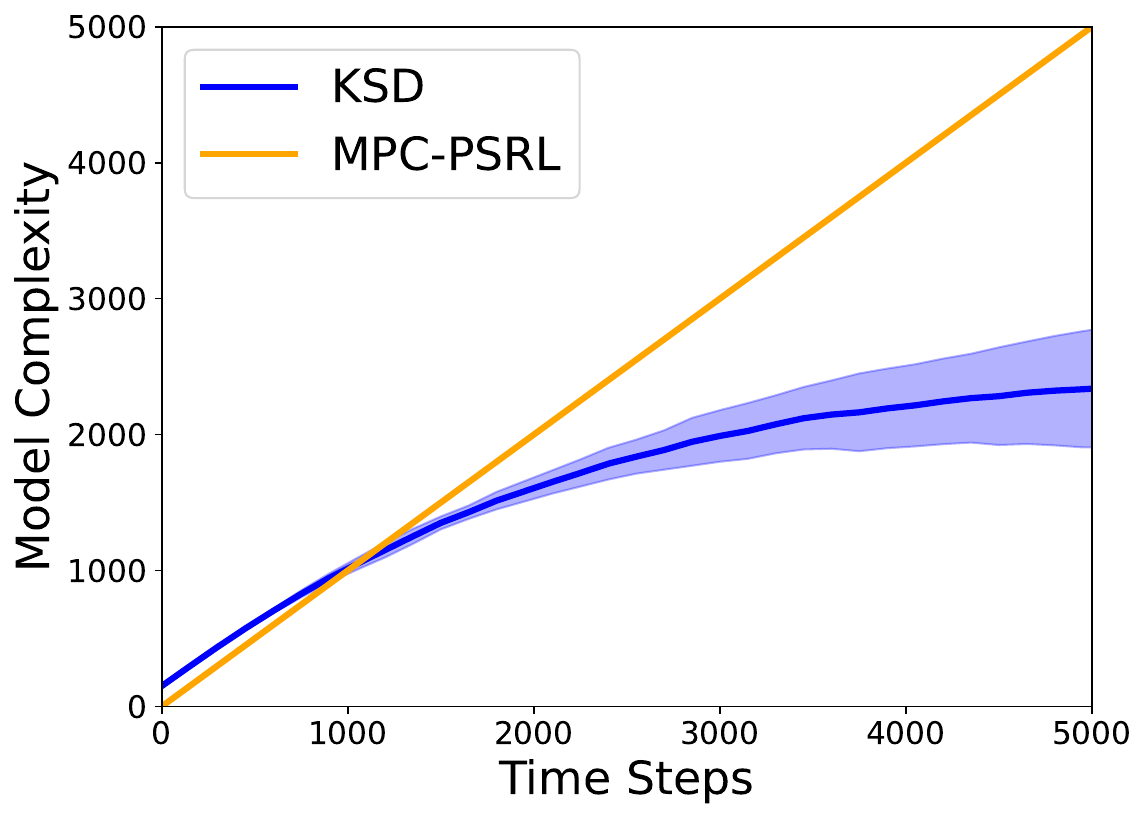}
         \caption{}
         \label{fig:reacher3}
     \end{subfigure}
%          \hfill
          \begin{subfigure}[b]{0.4\textwidth}
         \centering
\includegraphics[width=\textwidth]{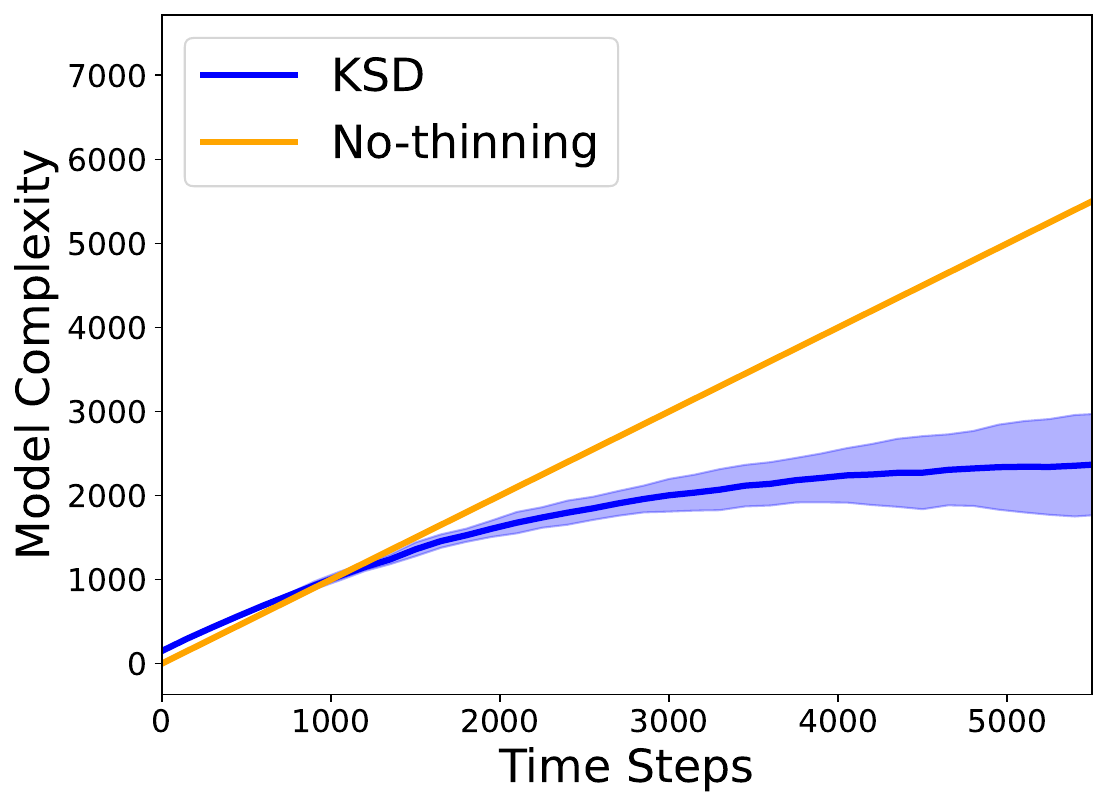}
         \caption{}
         \label{fig:reacher4}
     \end{subfigure}
        \caption{\textbf{(a)-(b)} compares the average cumulative reward return achieved by the proposed {\algo} (shown in blue) algorithm with MPC-PSRL \cite{pmlr-v139-fan21b}, SAC \cite{haarnoja2018soft}, and DDPG \cite{barth2018distributed} for \textbf{Reacher} with and without oracle rewards respectively. \textbf{(c)-(d)} compares the model-complexity.  We note that {\algo} is able to achieve the maximum average reward at-par with the current state of the art MPC-PSRL with drastically reduced model complexity. Solid curves represent the average across five trials (seeds), shaded areas correspond to the standard deviation amongst the trials}.
        \label{fig:reacher_plots}
\end{figure}

Using \eqref{integral_bound} into \eqref{here_1}, we get
\begin{align}
    \sum_{k=1}^{[\frac{T}{H}]}\mathbb{E}[{\Delta}_k(P)]  & \leq d \frac{2}{1-\alpha/2} T^{1-\frac{\alpha}{2}} R_{\max} H^{1+\frac{\alpha}{2}}.
\end{align}
The expression implies that
\begin{align}\label{final}
    \sum_{k=1}^{[\frac{T}{H}]}\mathbb{E}[{\Delta}_k(P)]  & = \mathcal{O}\left(d T^{1-\frac{\alpha}{2}} H^{1+\frac{\alpha}{2}}\right).
\end{align}
The same derivation would hold for the term $\mathbb{E}[{\Delta}_k(r)]$ (similar logic to \cite[Sec. 3.4]{pmlr-v139-fan21b}0), which would imply that $ \sum_{k=1}^{[\frac{T}{H}]}\mathbb{E}[{\Delta}_k(r)]   \leq \mathcal{O}\left(d T^{1-\frac{\alpha}{2}} H^{1+\frac{\alpha}{2}}\right)$ . From \eqref{refret_definition_theorem} and \eqref{final}, we can write
\begin{align}
    \sum_{k=1}^{[\frac{T}{H}]}\mathbb{E}[\Delta^{II}_k]  & = \mathcal{O}\left(dT^{1-\frac{\alpha}{2}} H^{1+\frac{\alpha}{2}}\right).
\end{align}
Hence proved. 
\end{proof}

\section{Additional Experiments and Analysis}\label{experimental_results}
In this section, we first provide details of the environments and the complexities added in order to validate multiple aspects of our \algo.
 
\textbf{Low-dimensional environments}: We consider the Continuous Cartpole ($d_s = 4$, $d_a = 1,H=200$) environment with a continuous action space which is a modified version of the discrete action classic Cartpole environment. The continuous action space enhances the complexity of the environment and makes it hard for the agent to learn. We also consider the Pendulum Swing Up ($d_s = 3$, $d_a = 1,H=200$) environment, a modified version of Pendulum where we limit the start state to make it harder and more challenging for exploration. To introduce stochasticity into the dynamics, we modify the physics of the environment with independent Gaussian noises ($\mathcal{N}(0,0.01)$). However, these environments are primarily lower dimensional environments. 

\textbf{Higher-dimensional environments} We consider the 7-DOF Reacher ($d_s=17,d_a=7,H=150$) and 7-DOF pusher ($d_s=20,d_a=7,H=150$) two challenging continuous control tasks as detailed in \cite{chua2018deep}. We increase the complexity from lower to higher dimensional environments and with added stochasticity to see the robustness of our algorithm and the consistency in performance. We conduct the experiments both with and without true oracle rewards and compare the performance with other baselines.

\begin{figure}
     \centering
     \begin{subfigure}[b]{0.32\textwidth}
         \centering
\includegraphics[width=\textwidth]{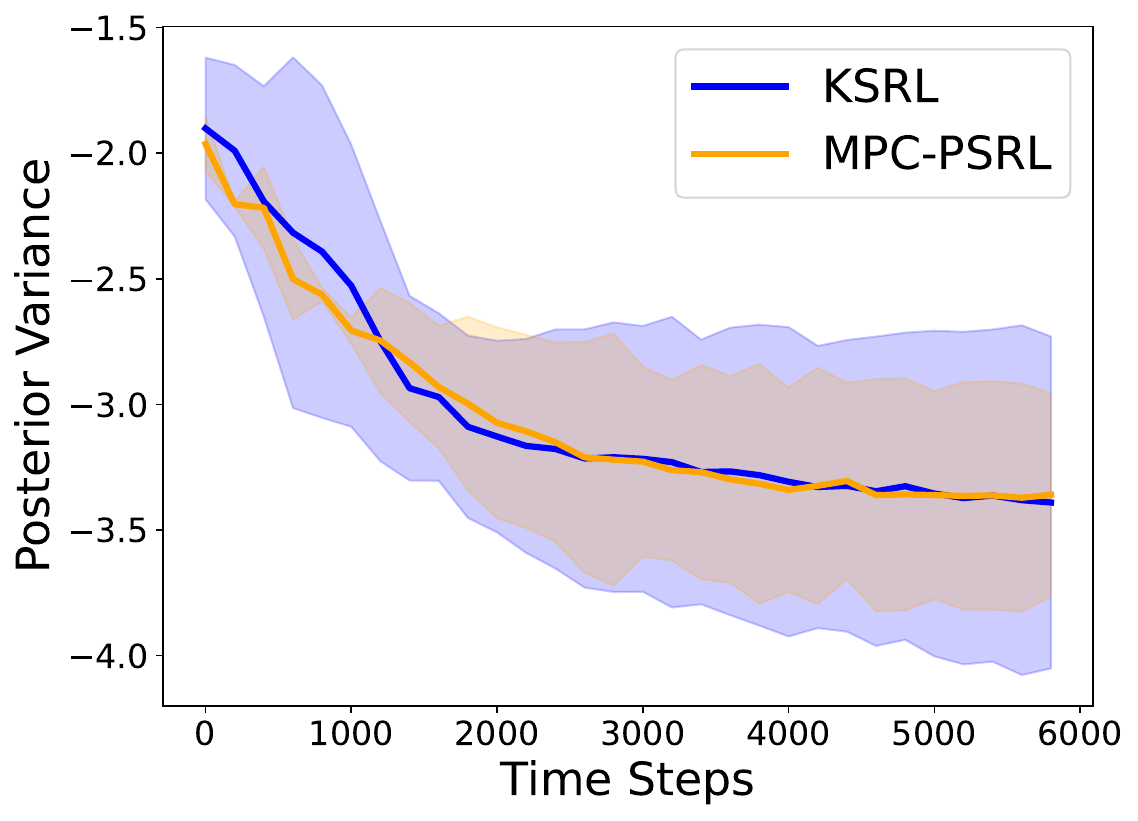}
         \caption{.}
         \label{fig:post_var1}
     \end{subfigure}
          \hfill
     \begin{subfigure}[b]{0.32\textwidth}
         \centering
\includegraphics[width=\textwidth]{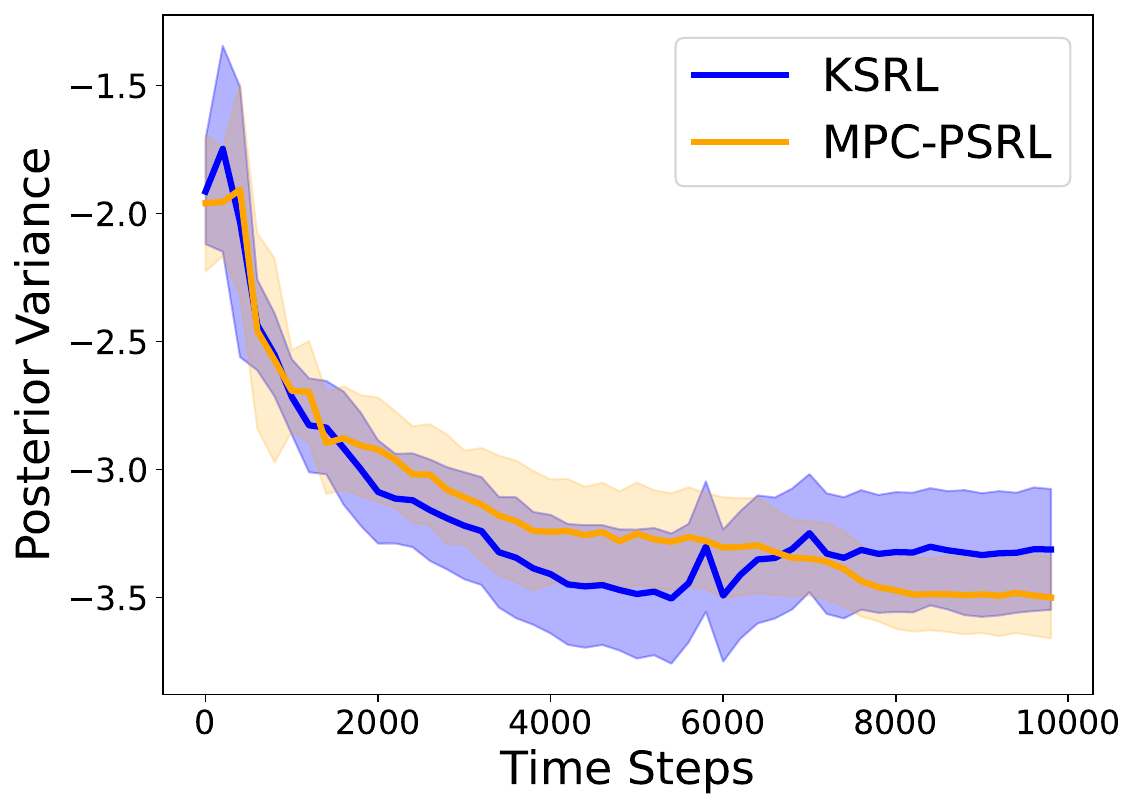}
         \caption{}
         \label{fig:post_var2}
     \end{subfigure}
     \hfill
          \begin{subfigure}[b]{0.32\textwidth}
         \centering
\includegraphics[width=\textwidth]{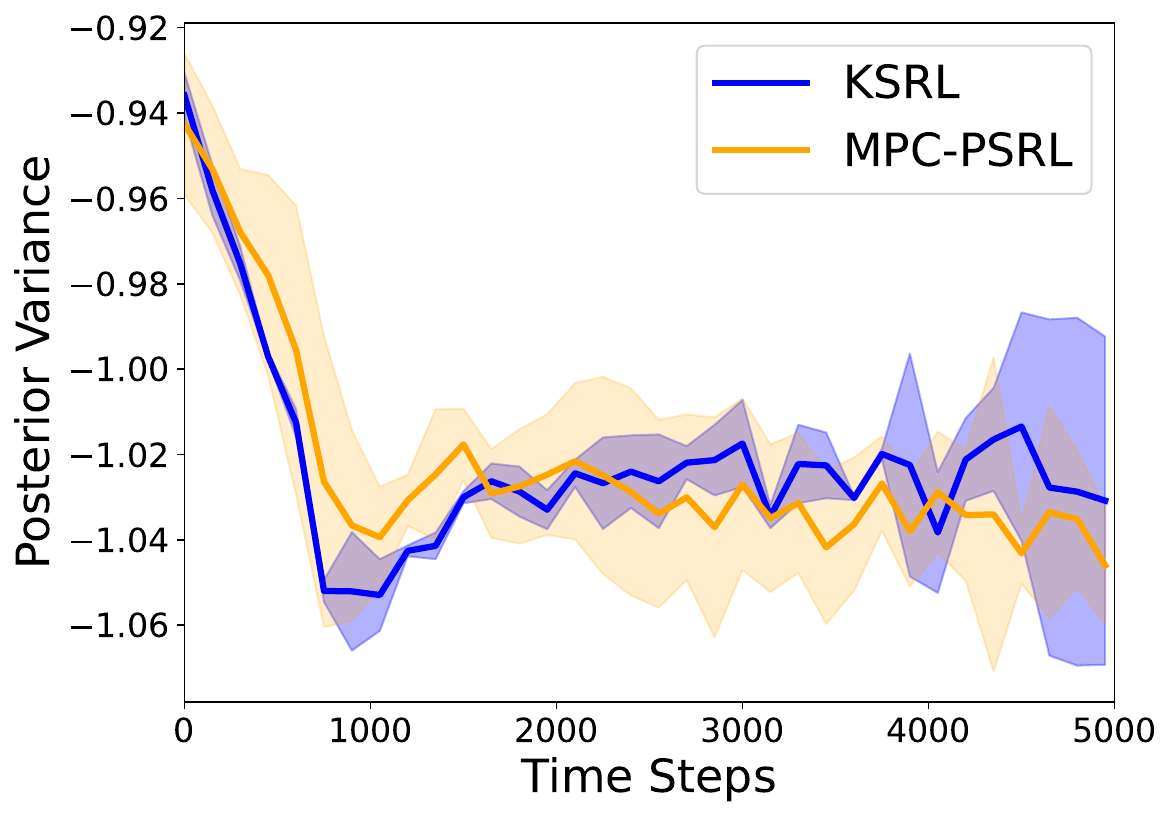}
         \caption{}
         \label{fig:post_var3}
     \end{subfigure}
        \caption{\textbf{(a)-(c)} compares the posterior variance of our \algo (blue) with MPC-PSRL \cite{pmlr-v139-fan21b} for Cartpole, Pendulum \& Reacher environments across the timesteps which shows we are learning the true posterior effectively without any significant bias (even we are compressing the dictionary). From \textbf{(a)-(c)} it is evident that posterior variance from \algo (blue) with compression converges to very similar posterior variance achieved by MPC-PSRL which highlights that the uncertainty estimation in our \algo is accurate. Plot is in logarithmic scale. Solid curves represent the average across five trials (seeds), shaded areas correspond to the standard deviation amongst the trials}
        \label{fig:post_var}
\end{figure}

\subsection{Experimental Details}
It is shown in literature \cite{pmlr-v139-fan21b} that a simple Bayesian Linear regression with non-linear feature representations learnt by training a Neural network works exceptionally well in the context of posterior sampling reinforcement learning. For a fair comparison of our \algo, we follow a similar architecture as \cite{pmlr-v139-fan21b} where we first train a deep neural network for both the transition and rewards model and extract the penultimate layer of the network for the Bayesian linear regression and Posterior sampling. As per our notation $s_i, a_i, s_{i+1}, r_i$ where $h_i = \ip{s_i, a_i}$ be the current state-action pair with the next state $s_{i+1}$ and reward $r_i$ and let's assume the representation from the penultimate layer of the deep neural network of the state-action pair is denoted by $z_i = \textit{NN}(h_i) \in R^d$ where $\textit{NN}$ is the trained deep neural network model and $d$ is the dimensionality of the representation. Then the Bayesian linear regression model deals with learning the posterior distribution $P_{\textit{post}} = P(\beta | D)$, with the linear model given by $\delta_i = \beta^T z_i + \epsilon$, where $\delta_i = s_{i+1} - s_i$ as suggested in \cite{deisenroth2013gaussian, pmlr-v139-fan21b} and $\epsilon \sim N(0, \sigma^2)$, $D$ is the size of the dictionary. Similar to prior approaches, we choose a multivariate Gaussian prior with zero mean and $\Sigma_{\textit{prior}}$ (conjugate prior) to obtain a closed-form estimation of the posterior distribution which is also multivariate Gaussian. Now, we sample $\beta$ from the posterior distribution $P_{\textit{post}}$ at the beginning of each episode and interact with the environment using MPC controller. As described in Appendix \ref{sec:Prelim} at each timepoint, the MPC applies the first action from the optimal action sequence under the estimated dynamics and reward function by solving $\arg\max_{a_{i:i+\tau}} \sum_{t=i}^{i+\tau} \mathbb{E}[r(s_{t}, a_{t})]$, where $\tau$ is the horizon and is considered as a hyperparameter. However, as described above the above method suffers from high computation complexity as the matrix multiplication step in posterior estimation is of order $O(d^2 N)$ which scales linearly with the size of the dictionary prior to that episode $N$. This not only enhances the computational complexity but also makes the optimization with MPC extremely hard. Hence, in our algorithm \algo we construct an efficient posterior coreset with Kernelized Stein discrepancy measure from \eqref{eq: stein kernel} and validate the average reward achieved with the compressed dictionary. We observe in all the cases with varied complexity, our algorithm \algo performs equally well or sometimes even better than the uncompressed current state of the art MPC-PSRL method with drastically reduced model complexity.

\subsection{Experimental Analysis}

We perform a detailed multifaceted analysis comparing our algorithm \algo with other baselines and state of the art methods in the continuous control environments. We perform the experiments in the environments with and without oracle rewards. The environmental setting of without oracle rewards is much more complex as here we have to model the reward function as well along with the dynamics which makes it much harder for the agent to learn with the added uncertainty in modelling. We also enhance the complexity of the environments by adding stochasticity which makes it harder for the agent to learn even for environments with oracle rewards and we also observe the performance by varying the dimensionality of the environments. In Figure \ref{fig:threegraphs_main_body} and Figure \ref{fig:with_rewards}, we compare our \algo with other baselines and SOTA algorithms for Cartpole, Pendulum and Pusher environments without and with oracle rewards respectively and Figure \ref{fig:reacher_plots} for Reacher with and without rewards. In all the cases, \algo shows performance at-par or even better for some cases with drastically reduced model complexity where we can see a benefit of $80\%$ improvement in the model complexity over the current SOTA with similar performance in terms of average rewards. In Figure \ref{fig:three graphs2}, we validate the average reward achieved by our \algo against baselines with respect to the runtime (wallclock time) in CPU minutes and clearly observe improved performance in-terms of wallclock time where \algo achieves optimal performance prior to the baselines and MPC-PSRL. We also perform the convergence analysis from an empirical perspective and observe the convergence in-terms of both Kernelized Stein Discrepancy and Posterior Variance. In Figure \ref{fig:three graphs2} \textbf{(d) -(f)}, we study the convergence in-terms of KSD and observe the convergence of our algorithm \algo without any bias and faster than the dense counterpart MPC-PSRL in-terms of wall clock time. We also study the convergence from the posterior variance perspective as it is extremely important for PSRL based algorithms to accurately estimate the uncertainty. Since, we are compressing the dictionary it is important to analyze and monitor the posterior variance over the timesteps to ensure that we are not diverging and close to the dense counterparts. In \ref{fig:post_var}, we observe the posterior variance achieved by \algo converges to the posterior variance achieved by MPC-PSRL \cite{pmlr-v139-fan21b} even though we are compressing the dictionary which highlights the efficacy of our posterior coreset. Finally, from the above plots and analysis we conclude that our our algorithm \algo achieves state of the art performance for continuous control environments with drastically reduced model complexity of its dense counterparts with theoretical guarantees of convergence.

\subsection{Details of Hyperparamters}
\begin{table}[H]
\centering
\begin{tabular}{|l|l|l|l|l|}
\hline
Environment                                                             & Cartpole & Pendulum & Pusher & Reacher \\ \hline
\begin{tabular}[c]{@{}l@{}}Steps\\ per episode\end{tabular} & 200      & 200      & 150    & 150     \\ \hline
Popsize                                                         & 500      & 100      & 500    & 400     \\ \hline
\begin{tabular}[c]{@{}l@{}}Number\\ of elites\end{tabular}      & 50       & 5        & 50     & 40      \\ \hline
\begin{tabular}[c]{@{}l@{}}Network \\ architecture\end{tabular} &
  \begin{tabular}[c]{@{}l@{}}MLP with \\ 2 hidden layers\\ of size 200\end{tabular} &
  \begin{tabular}[c]{@{}l@{}}MLP with\\ 2 hidden layers\\ of size 200\end{tabular} &
  \begin{tabular}[c]{@{}l@{}}MLP with\\ 4 hidden layers\\ of size 200\end{tabular} &
  \begin{tabular}[c]{@{}l@{}}MLP with\\ 4 hidden layers\\ of size 200\end{tabular} \\ \hline
\begin{tabular}[c]{@{}l@{}}Planning\\ horizon\end{tabular}      & 30       & 20       & 25     & 25      \\ \hline
Max iter                                                        & \multicolumn{4}{l|}{5}                 \\ \hline
\end{tabular}
\caption{Hyperparameters used for our algorithm \algo}
\end{table}

We keep the hyperparameters and the network architecture almost similar to \cite{pmlr-v139-fan21b} for a fair comparison.
For the baseline implementation of MPC-PSRL algorithm and our algorithm \algo we modify and leverage \footnote{\url{https://github.com/yingfan-bot/mbpsrl}} and observe that we were able to achieve performance as described in \cite{pmlr-v139-fan21b}. For obtaining the results from model-free algorithms as shown in  we use \footnote{\url{https://github.com/dongminlee94/deep_rl}}, and could replicate the results. Finally, for our posterior compression algorithm we modify the \footnote{\url{https://github.com/colehawkins/KSD-Thinning}}, \footnote{\url{https://github.com/wilson-ye-chen/stein_thinning}}  to fit to our scenario of posterior sampling reinforcement learning. We are thankful to all the authors for open-sourcing their repositories. Our implementation is available \footnote{\url{https://github.com/souradip-chakraborty/KSRL}}

\end{document}